\documentclass[runningheads]{llncs}

\usepackage[mobile]{eccv}

\usepackage{eccvabbrv}

\usepackage{graphicx}
\usepackage{booktabs}

\usepackage{caption}
\usepackage{subcaption}
\usepackage{wrapfig}
\usepackage[accsupp]{axessibility}  %

\usepackage{hyperref}

\usepackage{orcidlink}

\usepackage{xcolor}
\usepackage{xr}

\newcommand\NOTE[3]{\textcolor{#1}{[#2: #3]}}

\newcommand\laura[1]{\NOTE{olive}{Laura}{#1}}
\definecolor{jorange}{rgb}{8,.5,.0}
\newcommand\linus[1]{\NOTE{jorange}{Linus}{#1}}
\definecolor{bluegreen}{rgb}{.05,.6,.7}
\newcommand\marc[1]{\NOTE{magenta}{Marc}{#1}}
\newcommand\george[1]{\NOTE{green}{George}{#1}}
\definecolor{grey}{rgb}{.6,.6,.6}
\newcommand\peter[1]{\NOTE{blue}{Peter}{#1}}

 \renewcommand\marc[1]{}
 \renewcommand\linus[1]{}
 \renewcommand\laura[1]{}
 \renewcommand\george[1]{}
 \renewcommand\peter[1]{}

\usepackage{amsmath,amsfonts,bm}

\def\1{\bm{1}}

\def\eps{{\epsilon}}

\newcommand\arxivorelse[2]{#2} %

\DeclareMathAlphabet{\mathsfit}{\encodingdefault}{\sfdefault}{m}{sl}
\SetMathAlphabet{\mathsfit}{bold}{\encodingdefault}{\sfdefault}{bx}{n}

\def\gL{{\mathcal{L}}}

\def\Sgt{S}           %
\def\numimgs{N}            
\def\imgs{\{I_i\}^\numimgs_{i=1}}                %

\def\bb{\mathcal{G}}          %
\def\head{\mathcal{H}}          %
\def\headcam{\head_\textrm{cam}}          %
\def\headpts{\head_\textrm{pts}}          %
\def\headd{\head_\textrm{depth}}          %

\def\extr{\mathcal{E}}        %
\def\headthreed{\head_{3D}}          %

\def\d{d}                              %
\def\latent{z}                              %
\def\latenttwodlist{\{\latent^b_\textrm{2D}\}^B_{b=1}}               %
\def\latentthreedinit{\latent_\textrm{3D}}              %
\def\latentthreedpred{\hat{\latent}_\textrm{3D}}              %

\def\pointset{\textrm{X}}
\def\V{\Omega}                        %
\def\grid{\pointset_{\Omega}}        %
\def\eps{\epsilon}                    %
\def\x{\mathbf{x}}                      %
\def\xijk{\x_{i,j,k}}         %

\def\SBgt{\partial{S}}    %
\def\s{\mathbf{s}}                      %
\def\nearsurface{\pointset_\Sgt}

\def\dist{\mathrm{d}}

\def\sdfgt{f}                           %
\def\sdfpred{\hat{\sdfgt}}                           %
\def\gsdfgt{\sdfgt_{\grid}}                           %
\def\gsdfpred{\sdfpred_{\grid}}                 %
\def\maske{\textrm{M}_\textrm{eik}}                         %
\def\maskv{\textrm{M}_\textrm{val}} 

\def\Ltot{\mathcal{L}_{\mathrm{tot}}}     %

\def\lamc{\lambda_G}                  %
\def\lams{\lambda_s}                  %
\def\lamg{\lambda_\nabla}             %
\def\lame{\lambda_{\textrm{eik}}}     %

\renewcommand{\arxivorelse}[2]{#1}

\begin{document}

\title{Fus3D: Decoding Consolidated 3D Geometry from Feed-forward Geometry Transformer Latents}

\titlerunning{Fus3D}

\author{Laura Fink\inst{1}\orcidlink{0009-0007-8950-1790} \and
Linus Franke\inst{1,2}\orcidlink{0000-0001-8180-0963} \and
George Kopanas\inst{3}\orcidlink{0009-0002-5829-2192}
Marc Stamminger\inst{1}\orcidlink{0000-0001-8699-3442} \and
Peter Hedman$^\dagger$\orcidlink{0000-0002-2182-0185}
\vspace{-0.5em}
}

\authorrunning{L.~Fink et al.}

\institute{Friedrich-Alexander-Universität Erlangen-Nürnberg, Germany \and
Inria, Université Côte d'Azur, France\and
Google DeepMind, UK\\
}

\maketitle

\begingroup
\renewcommand\thefootnote{$\dagger$}
\footnotetext{Work done while at Google DeepMind, UK}
\endgroup
\vspace{-0.5em}

\begin{figure}
    \includegraphics[width=\textwidth]{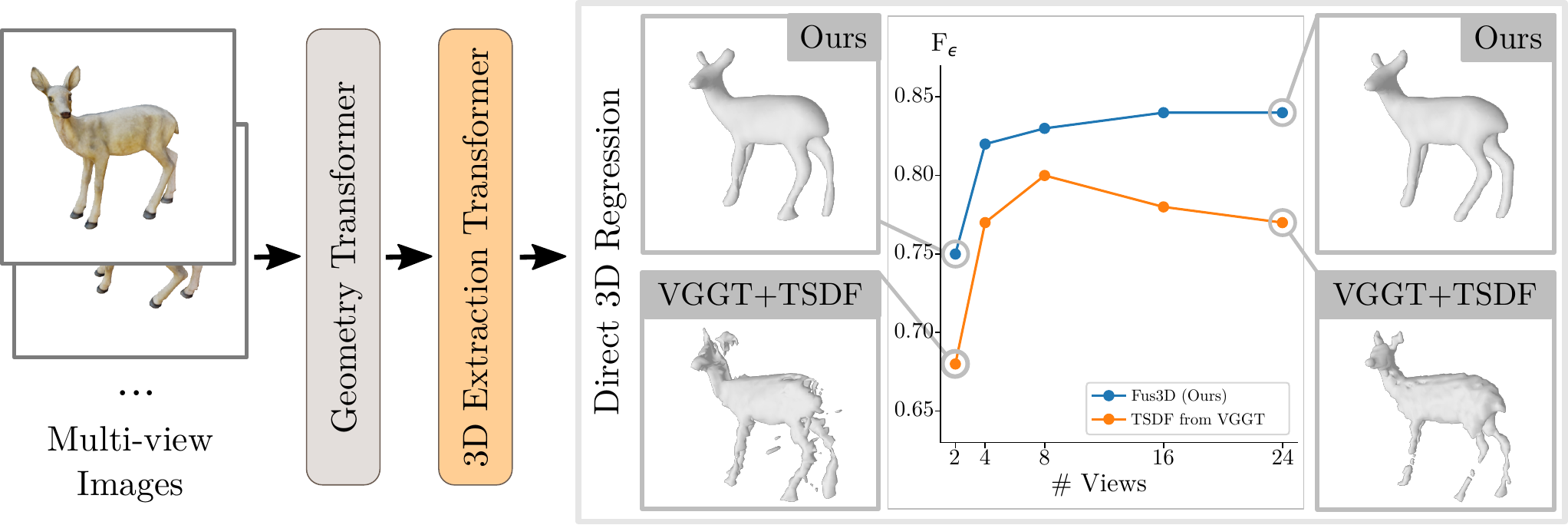}

    \vspace{-0.5em}
    \caption{
    Fus3D directly regresses a 3D representation from the latent space of a multi-view geometry transformer, bypassing per-view prediction and post-hoc fusion. 
    This, compared to VGGT~\cite{wang2025vggt} using TSDF fusion, yields improved surface completeness under sparse views (left, 2 views) and avoids error accumulation reducing details at scale (right, 24 views), also reflected in the F-score curves across view counts (center).
    }
    \label{fig:teaser}
\end{figure}

\vspace{-1em}
\begin{abstract}

We propose a feed-forward method for dense Signed Distance Field~(SDF) regression from unstructured image collections in less than three seconds, without camera calibration or post-hoc fusion. 
Our key insight is that the intermediate feature space of pretrained multi-view feed-forward geometry transformers~(FFGT) already encodes a powerful joint world representation; yet, existing pipelines discard it, routing features through per-view prediction heads before assembling 3D geometry post-hoc, which discards valuable completeness information and accumulates inaccuracies.

We instead perform 3D extraction directly from FFGT features via learned volumetric extraction: voxelized canonical embeddings that progressively absorb multi-view geometry information through interleaved cross- and self-attention into a structured volumetric latent grid. 
A simple convolutional decoder then maps this grid to a dense SDF. 
We additionally propose a scalable, validity-aware supervision scheme directly using SDFs derived from depth maps or 3D assets, tackling practical issues like non-watertight meshes.
Our approach yields complete and well-defined distance values across sparse- and dense-view settings and demonstrates geometrically plausible completions.

Code and further material can be found at {\small {https://lorafib.github.io/fus3d}}.
\end{abstract}

\section{Introduction}
\label{sec:intro}

3D reconstruction from unstructured image collections and video remains a fundamental challenge in computer vision, with broad implications for downstream tasks in semantic understanding, robotics, and scene interaction.
A key requirement in such settings is a complete 3D representation, e.g. when accurate distance information is critical.
Distance information, often in the form of Signed Distance Functions (SDFs), is difficult to acquire when input views are sparse or otherwise poorly conditioned.
In this work, we provide a novel method for this online and sparse-view setting by leveraging extensively pretrained multi-view geometry transformers and directly regressing 3D SDFs from them.

Recent feed-forward multi-view geometry transformers, such as DUSt3R~\cite{wang2024dust3r}, VGGT~\cite{wang2025vggt}, or DA3~\cite{lin2025depthanything3}, have demonstrated remarkable capabilities in recovering scene geometry directly from images, producing per-image depth or point maps alongside camera poses and intrinsics in a single forward pass. 
Central to their success is a powerful geometry prior acquired through large-scale training, enabling robust reconstruction even under challenging conditions such as strong symmetry or minimal inter-view overlap -- scenarios that confound classical feature-matching pipelines.

Despite these advances, a fundamental bottleneck remains in how 3D scenes are ultimately assembled from the per-image predictions of these transformers. 
Current pipelines rely on classical post-hoc fusion of per-view outputs, typically via point cloud merging, Truncated SDF (TSDF) fusion or Poisson reconstruction to obtain a coherent 3D representation. 
This design has two consequential failure modes.
First, in sparse-view settings, the transformer's learned world prior is only leveraged locally, per image, leaving unobserved regions geometrically underspecified and yielding incomplete reconstructions (Fig.~\ref{fig:teaser}, bottom left). 
Second, when processing many views, minor per-view inconsistencies accumulate and propagate into the fused representation, introducing noise that degrades reconstruction fidelity at scale (Fig.~\ref{fig:teaser}, bottom right, red curve).

We address both failure modes through a single unifying idea: rather than fusing per-view outputs in 3D space, we directly regress the SDF from the transformer's intermediate latent features. 
This allows the full multi-view geometry prior, jointly built up across all input views, to inform the 3D estimate at every spatial location, yielding complete reconstructions in sparse-view regimes and noise-free scaling with more views (see crops in Fig.~\ref{fig:teaser}).

The central technical challenge is the dimensionality mismatch between the set of 2D input tokens and the 3D target representation. 
We resolve this by introducing a learned embedding query that attend into the geometry transformer's feature space, effectively lifting 2D multi-view features into a 3D structured latent space without requiring 
any 3D-to-3D autoencoder training regimes. 
Crucially, we perform repeated extraction during the feature construction process of the multi-view transformer in order to build meaningful 3D latent features. 
As we demonstrate, the resulting feature space encodes a rich geometry prior including object-level symmetry and shape statistics which enables geometrically complete and well-defined SDFs even for unobserved surfaces.
Finally, the intermediate 3D feature volume reveals emergent semantic structure that could serve as a foundation for downstream perception.

In summary, we make the following contributions:
\begin{itemize}
    \item We identify the per-view prediction bottleneck in feed-forward geometry transformers as the root cause of both sparse-view incompleteness and noise accumulation under many views, and thus propose 3D extraction directly from the transformer's intermediate feature space.
    \item We introduce a learned volumetric extraction module: a set of 3D-position-conditioned canonical embeddings that progressively fuse multi-view geometry features into a structured latent grid via 
    attention, enabling projection-free 2D-to-3D lifting. 
    \item We propose a validity-aware SDF supervision scheme, 
    to handle non-watertight meshes and unobserved regions gracefully, enabling scalable training on large-scale real-world datasets.
    \item Our results indicate a geometric shape prior encoded within the 3D feature space,
    enabling complete and robust reconstructions in challenging scenarios.
\end{itemize}

\section{Related Work}
\label{sec:related}

\paragraph{Classical and Online Reconstruction.}
Classical pipelines using Structure-from-Motion~(SfM) and Multi View Stereo~(MVS) remain strong baselines for reconstruction under sufficient overlap and texture~\cite{schonberger2016pixelwiseview, sfm,seitz2006comparisonevaluation}, while SLAM systems coupled with TSDF fusion enable real-time incremental reconstruction in RGB-D settings~\cite{millane2024nvblox, niessner2013real,mascaro2025scenerobots, sayed2022simplerecon}. 
Both approaches, however, handle camera calibration, depth estimation and geometry fusion as distinct tasks.

\paragraph{Analysis-by-Synthesis}
Neural rendering methods formulate 3D reconstruction through view synthesis based on various scene representations, with NeRF~\cite{ mildenhall2021nerfrepresenting} and Gaussian splatting~\cite{kerbl20233dgaussian} as canonical examples.
Learning-based implicit shape representations such as SDFs and occupancy functions have become popular choices for geometry modeling~\cite{park2019deepsdf, mescheder2019occupancy}.
NeuS and its follow-ups achieve high-fidelity surface 
reconstruction by coupling SDF representations with differentiable 
rendering~\cite{wang2021neuslearning, li2023neuralangelohigh, neus2}.

\paragraph{Feed-Forward Geometry from Unposed Images.}
DUSt3R jointly recovers dense geometry and camera parameters from unconstrained image collections via point-map regression~\cite{wang2024dust3r}, substantially reducing the separation between matching, pose estimation, and dense reconstruction. 
VGGT, $\pi^3$, and MapAnything extend this trend to broader geometric outputs, in the form of depth, point maps, cameras, and tracks, with strong generalization across variable numbers of views~\cite{wang2025vggt, wang2025pi3,keetha2026mapanything}. 
While VGGT achieves impressive geometric generalization, its outputs remain per-view predictions that must be consolidated via TSDF fusion to yield a surface, discarding the joint multi-view prior in the process; this predict-then-fuse pipeline serves as a direct baseline for our approach.

A parallel line of work augments feed-forward prediction with persistent spatial memory: Spann3R maintains a global-frame memory for consistent point map prediction~\cite{wang2025spann3r}, while CUT3R develops a stateful formulation for improved temporal consistency~\cite{wang2025cut3r}. 
Despite impressive progress, all of these methods produce per-pixel or per-patch 2.5D outputs and require post-hoc fusion to assemble coherent 3D geometry. 
Their proposed spatial memories support only ray-map or pixel-aligned readouts~\cite{wang2025cut3r}, not direct 3D positional queries. 
Our work departs from this paradigm entirely, performing 3D extraction directly in the transformer's latent space without any per-view prediction or post-hoc fusion.

\paragraph{Generalizable Reconstruction via Differentiable Rendering.}
Generalizable novel view synthesis~(NVS) methods reconstruct radiance fields or Gaussian primitives from sparse views, under known~\cite{chen2024mvsplat, charatan2024pixelsplat, zhang2024transplatgeneralizable} or unposed~\cite{jiang2025anysplat, xu2024grm} cameras. 
Generalizable reconstruction methods such as LRM~\cite{hong2024lrmlarge} and its follow-ups~\cite{he2023openlrmopen, zhuang2024gtr} regress structured 3D representations from sparse images via large transformers, with MeshLRM~\cite{wei2025meshlrmlarge} extending this to end-to-end mesh reconstruction through differentiable mesh rendering.
M-LRM~\cite{li2025multi} proposes 3D aware positional encoding for latent features projected into voxel grids, and GeoLRM~\cite{zhang2024geolrm} fuse voxel-based sparse latent features to decode Gaussians, showcase the promising direction of 3D aware reconstruction in posed scenarios.

Post-hoc consolidation methods address view-aligned primitive redundancy via fusion~\cite{jiang2025anysplat}, pruning~\cite{ye2025yonosplat}, or projecting latents via depth maps into grids~\cite{wang2025volsplat}, but remain bound to the predict-then-fuse composition. 
We instead aggregate multi-view features directly into a 3D latent volume, avoiding view-aligned intermediate predictions entirely.
Generalizable SDF regression approaches, such as SparseNeuS~\cite{long2022sparseneusfast}, VolRecon~\cite{ren2023volreconvolume}, and UFORecon~\cite{na2024uforecongeneralizable} use  learned correspondence finding based on cost volumes or frustum features and neural SDF decoding.
Hence, VolRecon and UFORecon represent the closest prior work to ours in terms of output representation and serve as our primary geometry quality baselines; however, both require known camera parameters at test time 
in contrast to our fully feed-forward, pose-free formulation.

\paragraph{3D Generative Methods.}
Autoencoder-based pipelines compress explicit 3D representations into volumetric latent codes that support generation, interpolation, and image-conditioned reconstruction~\cite{trellis, cheng2023sdfusion, chang2025reconviagen, mittal2022autosdf,gao2025can3tok}. 
These structured latent spaces have proven especially expressive for high-fidelity geometry generation and are increasingly paired with diffusion models for diverse shape synthesis~\cite{trellis}. 
Our extraction transformer shares this volumetric latent structure but departs fundamentally in how it is trained: autoencoder pipelines require 3D-to-3D training in order to learn a meaningful latent representation, thus relying on high quality 3D data not only as supervision signal but also as input to the encoder which is often omitted for the actual generation task~\cite{trellis}.
While direct 3D supervision is replaceable using differential rendering, the reliance on 3D data as input during training creates a bottleneck for dataset generation~\cite{jiang2025rayzer}. 
By leveraging a pretrained backbone with 2D-to-3D training, we shift the burden of representation learning to scalable 2D models, using limited 3D data only to lift their powerful priors into a consolidated 3D space.

\section{Method}
\label{sec:method}
Contemporary geometry transformers~$\bb$~\cite{wang2024dust3r, wang2025vggt} encode a set of context images $\imgs$ into per-patch features, processed by interleaved frame-local self-attention and frame-global cross-attention blocks before being routed to specialized 2D decoder heads $\headd, \headpts, \headcam$ that produce per-image depth maps, point maps, and camera parameters in one-to-one correspondence with input patches. A coherent 3D scene is then recovered via post-hoc fusion, lifting per-view predictions into a shared coordinate frame via point cloud followed by a surface reconstruction step. This is entirely decoupled from the transformer's internal representations and misses the opportunity to leverage the rich priors encoded in the model. 
In this design, the expressive joint multi-view feature space assembled by $\bb$ is discarded before the 3D representation is constructed. 

\begin{figure}[t]     
    \centering
    \includegraphics[width=1\linewidth]{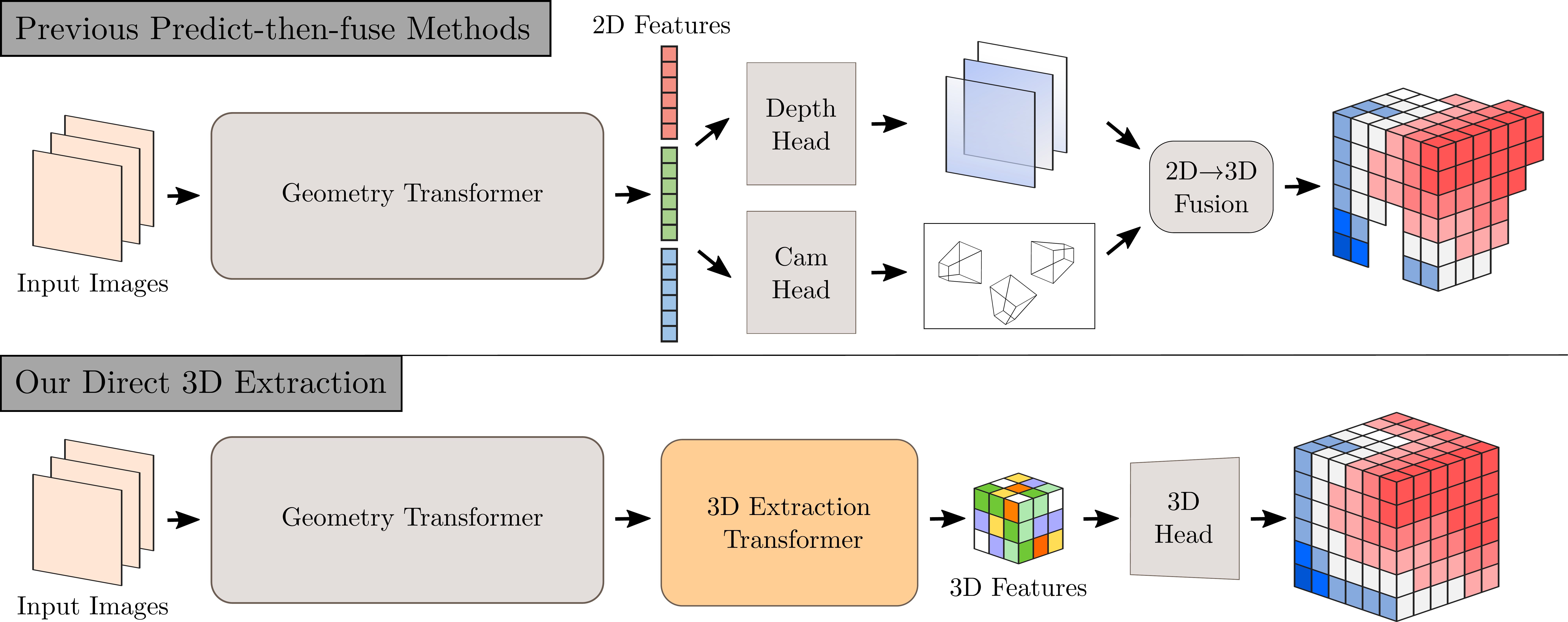}
    \caption{
    Juxtaposition of geometric extraction metholodogies. Common pipelines like VGGT~\cite{wang2025vggt} (top) 
    route transformer features through per-view 2D decoder heads, discarding the joint multi-view representation before 3D assembly.
    We instead extract dense 3D features directly from the transformer's intermediate feature space, preserving the full multi-view information.
    }
    \label{fig:extraxtion_2Dvs3D}
    
\end{figure}

In contrast, we replace the per-view decoder heads with a learned 3D extraction module $\extr$ that operates directly on the transformer's intermediate feature space. Rather than projecting back to individual image planes, $\extr$ aggregates multi-view geometry features into a dense volumetric feature grid via attention, without explicit camera projection or per-view fusion, allowing the full joint multi-view prior to inform every spatial location simultaneously.
A juxtaposition of these two paradigms can be seen in Fig.~\ref{fig:extraxtion_2Dvs3D}.

\subsection{Learned Volumetric Extraction from Multi-View Features}

\begin{figure}[h]     
    \centering
    \includegraphics[width=1\linewidth]{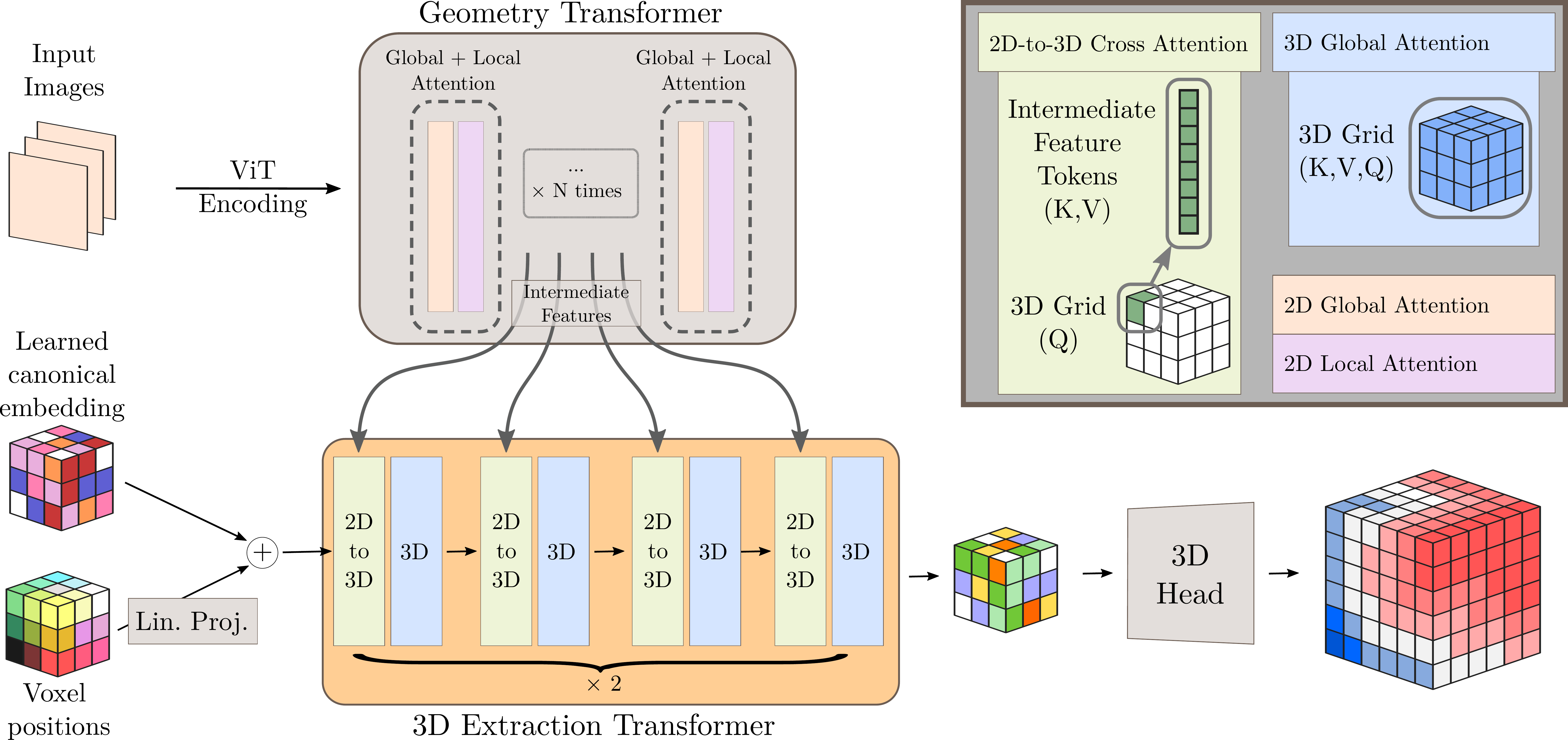}
    \caption{Architecture of Fus3D: The geometry transformer~$\bb$ (beige) processes tokenized input images, yielding a list of 2D intermediate features~$\latenttwodlist$ extracted from different stages. 
    The extraction transformer~$\extr$ (orange) leverages 2D-to-3D cross attention (green) to absorb the 3D information into features of the learned canonical embedding~$\latentthreedinit$, and distributes this information via 3D self-attention (blue) throughout the volumetric latent.
    The head~$\headthreed$ decodes the resulting structured latent~$\latentthreedpred$ into a dense SDF grid.}
    \label{fig:ff3d_sdf}
\end{figure}

To directly regress a 3D representation from a set of input images $\imgs$, we propose a three-stage pipeline: a pretrained geometry transformer backbone~$\bb$ that provides powerful multi-view geometry features; a 3D extraction transformer~$\extr$ that fuses the intermediate features of $\bb$ into a dense volumetric latent grid; and a 3D decoder head~$\headthreed$ that maps these latent features to a dense grid of SDF values~$\gsdfpred$ in a grid $\grid$ over a volume domain $\V$. 
An overview of the architecture is given in Fig.~\ref{fig:ff3d_sdf}.

\paragraph{Backbone.}
Our design requires that all input views are processed within a unified feature space, such that intermediate representations are directly comparable across frames. 
This property holds for geometry transformers like VGGT~\cite{wang2025vggt}, DA3~\cite{lin2025depthanything3}, and $\pi^3$~\cite{wang2025pi3}, which process all views jointly in a shared token sequence, but not for pair-wise methods such as MASt3R~\cite{leroy2024mast3r} or DUSt3R~\cite{wang2024dust3r}, which handle image pairs independently and aggregate predictions post-hoc. 
We therefore adopt the transformer architecture of VGGT as the geometry backbone~$\bb$. 
As a direct consequence, no explicit conditioning on precomputed camera calibration is required. 
We retain VGGT's coordinate convention, placing the scene's coordinate frame at the camera center of the first input view.

\paragraph{Learned Canonical Embedding.}
The mismatch in dimensionality between per-image feature representations and the target 3D output is a major difficulty for 3D feature extraction from 2D latents from $\bb$.
To address this problem, we take inspiration from neural memory approaches for continuous reconstruction~\cite{wang2025spann3r,wang2025cut3r}, in which a scene-agnostic learned embedding, initialized independently of any observed image, serves as the initial state of a memory that is subsequently updated through attention. 
We generalize this idea to 3D: rather than using a 2D positional encoding, we condition a set of learned embeddings~$\latentthreedinit$ (we use extents of $16^3$ and dimension $\d{=}2048$) on their corresponding 3D voxel positions within the volume~$\V$. 
These position-conditioned latent features constitute the initial state of the 3D memory, into which multi-view geometry information is progressively fused and compressed.

The voxel-grid structure implies that the region of interest must be approximately known at inference time. 
While the bounding box of $\V$ can, in principle, vary per scene, we adopt a scheme whereby the datasets are considered in a normalized coordinate space, yielding a fixed bounding box relative to the first input view, common in the field~\cite{ding2022transmvsnetglobal, yao2018mvsnetdepth, ren2023volreconvolume, na2024uforecongeneralizable}.

\paragraph{Multi-view Feature Aggregation.}
Following the design of 2D decoder heads~\cite{wang2025vggt}, we adopt a hierarchical design to compute volumetric features from our learned 3D embeddings $\latentthreedinit$ and the intermediate 2D latent features~$\latenttwodlist$ of the backbone~$\bb$.
Our 3D extraction transformer~$\extr$ lifts and computes the updated volumetric \textit{3D} features as
{\small
\begin{equation}
    \latentthreedpred = \extr\!\left(\latentthreedinit,\, \latenttwodlist\right),
\end{equation}
}
elements of $\latenttwodlist$ are injected at distinct stages of $\extr$, as seen in Fig.~\ref{fig:ff3d_sdf}.

Concretely, the embeddings~$\latentthreedinit$ serve as the initial queries for the first cross-attention module of~$\extr$, whose keys and values are drawn from the first intermediate feature map~$\latent^1_\textrm{2D}$. 
The attended volume is then passed through a volumetric self-attention block, which propagates information across spatial locations in the volume~$\V$. 
Following the interleaved local--global attention design of Wang et al.~\cite{wang2025vggt}, these two operations form a single transformer block. 
The process is repeated for each intermediate feature map in~$\latenttwodlist$ ($B{=}4$) with the updated 3D features~$\latentthreedpred$ carried forward between stages; the full four-stage sequence is then applied twice, yielding $8$ transformer blocks in total.

\subsection{Feed-forward SDF Regression}
\label{sec:ffsdf}

We define the ideal signed distance function $\sdfgt$ for the shape~$\Sgt \subset \mathbb{R}^3$ as
{\small
\begin{equation}
\sdfgt(\x) =
\begin{cases}
-\dist(\x, \SBgt), & \x\in \Sgt,\\[1mm]
\phantom{-}\dist(\x, \SBgt), & \x\notin\Sgt,\\
\end{cases}
\end{equation}
}
where $\x \in \Sgt$ are points inside the shape,
$\x \notin \Sgt$ are outside and  $\x \in \SBgt$ are at the boundary of $\Sgt$.
The unsigned distance is then $\dist(\x, \SBgt) = \min_{\s\in\SBgt} ||\x-\s||_2.$

\paragraph{Decoding.}

The volumetric latent features~$\latentthreedpred$ produced by $\extr$ are decoded by $\gsdfpred = \headthreed(\latentthreedpred)$ with a resolution of $64^3$.
$\headthreed$ is a simple convolutional upsampling network. 
$\gsdfpred$ represents the SDF values $\sdfpred(\xijk)$ corresponding to the voxel center positions~$\xijk$ at index $(i,j,k)$ of grid~$\grid$ 
sampling the bounding volume~$\V$. The voxel size is denoted as $\eps$.

\paragraph{Supervision.}
Following prior SDF works~\cite{yariv2024mosaicsdf, yariv2020multiviewneural}, we supervise the 
network with a combination of a direct value loss $\gL_\textrm{SDF}$, a gradient loss $\gL_{\nabla}$, and add an Eikonal regularizer $\gL_{\textrm{eik}}$.
However, in practice, target SDF values may not be well-defined everywhere within~$\V$, requiring masking.

\paragraph{Validity Masks.}
TSDF fusion leaves unobserved regions without valid values, and large-scale 3D asset datasets (Sec.~\ref{sec:datasets}) frequently contain non-watertight meshes, for which the interior is ambiguous and spurious sign flips arise in case of open boundaries (see \arxivorelse{App. Fig.~\ref{fig:masking}}{the supplemental material}).
Rather than relying on mesh preprocessing~\cite{li2025sparc3d, yariv2024mosaicsdf}, we instead model these failure modes in our training loss through two binary masks defined on $\grid$.

The \emph{validity mask}~$\maskv \subseteq \grid$ identifies positions~$\x$ for which the ground-truth SDF value is well-defined and reliable; all loss terms are evaluated exclusively on $\pointset \subseteq \maskv$.

The \emph{Eikonal mask}~$\maske \subseteq \grid$ identifies positions in which the ground-truth SDF is locally smooth and sign-consistent. 
It is constructed by evaluating $\gL_{\textrm{eik}}$ on the ground-truth field~$\gsdfgt$ itself. 
If the Eikonal term is greater than 1 by a sufficiently large amount, we classify the value and its local neighborhood as unreliable. 
Consequently, unreliable regions are excluded from sign-sensitive supervision.
Using these masks, we adapt the regression losses in the following ways:
{\small
\begin{align}
    \gL_\textrm{SDF}(\pointset) &= 
        \frac{1}{|\pointset|}\sum_{\x_n \in \pointset}
        \begin{cases}
            \left\| \sdfpred(\x_n) - \sdfgt(\x_n)\right\|_1, 
                & \x_n\in \maske,\\[2mm]
            \left\|\, |\sdfpred(\x_n)| - |\sdfgt(\x_n)|\,\right\|_1, 
                & \x_n\notin \maske,
        \end{cases}\\[6pt]
    \gL_{\nabla}(\pointset) &= 
        \frac{1}{|\pointset|}\sum_{\x_n \in \pointset}
        \begin{cases}
            \left\| \nabla_{\x}\sdfpred(\x_n) - \nabla_{\x}\sdfgt(\x_n) \right\|_2,
                & \x_n \in \maske,\\[2mm]
            0, & \x_n \notin \maske.
        \end{cases}
\end{align}
}
Outside $\maske$, $\gL_\textrm{SDF}$ degrades gracefully to an unsigned distance 
variant, preserving a meaningful signal without imposing incorrect sign supervision.

Additionally, we keep the camera loss $\gL_\textrm{cam}$ by Wang et al.~\cite{wang2025vggt} to retain the $\bb$'s camera estimation functionality.

\paragraph{Total Loss.}
We combine the above terms as:
{\small
\begin{equation}
\Ltot
=
\lams\, \gL_\textrm{SDF}(\nearsurface)
+
\lamc\, \gL_\textrm{SDF}(\grid)
+
\lamg\, \gL_{\nabla}(\grid)
+
\lame\, \gL_{\textrm{eik}}(\grid)
+
\gL_\textrm{cam},
\end{equation}
}
where $\gL(\grid)$ refer to losses evaluated on voxel centers of grid~$\grid$, and 
additionally employ $\gL_\textrm{SDF}(\nearsurface)$, where $\nearsurface$ refers to sample positions close to the the shape boundary~$\SBgt$ to put extra emphasis on surface extraction.
The gradients for $\gL_{\nabla}(\grid)$ and $\gL_{\textrm{eik}}(\grid)$ are computed by finite differences on the grid.

\section{Experimental Setup} 
\label{sec:setup}

\subsubsection{Datasets and Sampling\\}
\label{sec:datasets}

\hspace{-0.5em}All experiments require scenes annotated with images, camera parameters, and a surface mesh. 
Meshes are converted to SDFs or TSDFs~\cite{millane2024nvblox} via rendered depth maps on the fly (see \arxivorelse{App. Sec.~\ref{supp:sdf_supervision}}{the supplemental material} for details). Unlike VGGT~\cite{wang2025vggt}, we do not normalize scale relative to the first view's depth, instead normalizing to a dataset-global scene range. 
Both datasets feature restricted camera setups, simplifying the regression task (see \arxivorelse{App. Sec.~\ref{supp:dataset_discussion}}{the supplemental material}).
Mini-batches of size $b$ are constructed by drawing $b$ scenes, each with 2--8 unique views sampled uniformly without overlap constraints. 
Images are rescaled to 448\,px on the longer side; augmentation crops the shorter side to 228--448\,px and applies consistent color jitter on each view set.

\paragraph{Objaverse.}
We use ${\approx}5\%$ of the \textsc{Objaverse}~\cite{objaverse} subset used by Xiang 
et al.~\cite{trellis}: 17K object-centric scenes with 24 views each, and 170 
held-out scenes for evaluation. Supervision meshes are carved to retain only 
triangles potentially visible from outside the object. We note that this is a 
comparably small training set, leaving substantial headroom for improvement at 
larger scale.

\paragraph{DTU.}
For forward-facing \textsc{DTU} scenes~\cite{dtu}, objects are only partially observed, so the ground-truth mesh is open and SDF signs (inside/outside) are ill-defined.
Hence, we adapt the calculation of losses in two ways: 
we compute $\gL_\textrm{SDF}(\grid)$ from a depth-fused TSDF. 
For $\gL_\textrm{SDF}(\nearsurface)$, we keep using the GT mesh but resort to its unsigned variant by setting $\maske$ to the empty set during its computation.
The setup of train and test splits follows VolRecon~\cite{ren2023volreconvolume} and 
UFORecon~\cite{na2024uforecongeneralizable}.

\subsubsection{Training\\}
\label{sec:training}

\hspace{-1.6mm}\textit{Initialization.}
The backbone~$\bb$ is initialized with the publicly available VGGT weights~\cite{wang2025vggt_checkpoint}, which use DINOv2 for patch tokenization~\cite{oquab2024dinov2learning}. 
The 3D decoder head~$\headthreed$ follows the structure generation decoder of Xiang et al.~\cite{trellis} with the final sigmoid activation removed, initialized from their released weights accordingly. 
The extraction transformer~$\extr$ and canonical embedding~$\latentthreedinit$ are trained from scratch.

\paragraph{Backbone finetuning.}

In these experiments, we specialize our back-end to work better for object-centric datasets. Inspired by Ren et al.~\cite{ren2025fin3r}, we achieve this through light-weight LoRA fine-tuning~\cite{hu2022lora} and apply the same fine-tuning strategy consistently across all comparison methods.

\paragraph{Training Curriculum.}
We employ a two-stage curriculum. In the first stage, the SDF is predicted at low resolution ($16^3$) using a single linear decoder, supervised with $\gL_\textrm{SDF}$ and $\gL_\textrm{cam}$ only. 
In the second stage, $\headthreed$ is introduced to predict the full-resolution SDF grid under the complete loss described in Sec.~\ref{sec:ffsdf}. 
For \textsc{DTU} experiments, we fine-tune the resulting weights on the \textsc{DTU} training set.

Following related work~\cite{jiang2025anysplat}, we use Adam with decoupled weight decay regularization as optimizer~\cite{loshchilov2017decoupled} and a cosine learning rate scheduler with warm up for all stages.
Training uses up to four NVIDIA A100 80\,GB GPUs.

\section{Results}
\label{sec:results}

We assess the quality of our method’s predictions in two complementary settings: With generalizable implicit methods and with pose-free feed-forward fusion baselines. We further provide analysis to validate our key design decisions.

\subsubsection{Comparison with Feed-Forward SDF Baselines\\}
\label{subsec:eval_dtu}

\hspace{-0.6em}We compare with VolRecon~\cite{ren2023volreconvolume} and UFORecon~\cite{na2024uforecongeneralizable} on the \textsc{DTU} dataset.
Given the domain gap between \textsc{Objaverse}, we finetune on their proposed training split (see Sec.~\ref{sec:setup}).
Note that this fine-tuning reduces the geometric extrapolation, so that it is limited to potentially visible regions for which a supervision signal is available.
Since both methods require known camera parameters, we supply VGGT-predicted poses and intrinsics in place of ground-truth, placing all methods in a fully feed-forward, pose-free setting consistent with our own. 
We note that this may disadvantage the baselines, which were designed for precise camera input; for completeness, additional comparisons are shown in \arxivorelse{App. Sec.~\ref{supp:eval_dtu}}{the supplemental material}, supplying VolRecon and UFORecon with COLMAP-estimated and ground-truth poses, though these settings forfeit the feed-forward assumption and introduce additional computational overhead.
Coordinate systems from different modalities are aligned to the ground-truth frame utilizing standard alignment procedures~\cite{besl1992icp, open3d, knapitsch2017tankstemples}; see \arxivorelse{App. Sec.~\ref{supp:alignment}}{the supplemental material} for details.
In line with UFORecon~\cite{na2024uforecongeneralizable}, we distinguish between sets with substantial view overlap as \emph{favorable} configurations, and \emph{unfavorable} ones, where views are far apart with limited shared visibility, posing a harder problem.

\begin{figure}[htb]
    \centering
    \includegraphics[width=\textwidth]{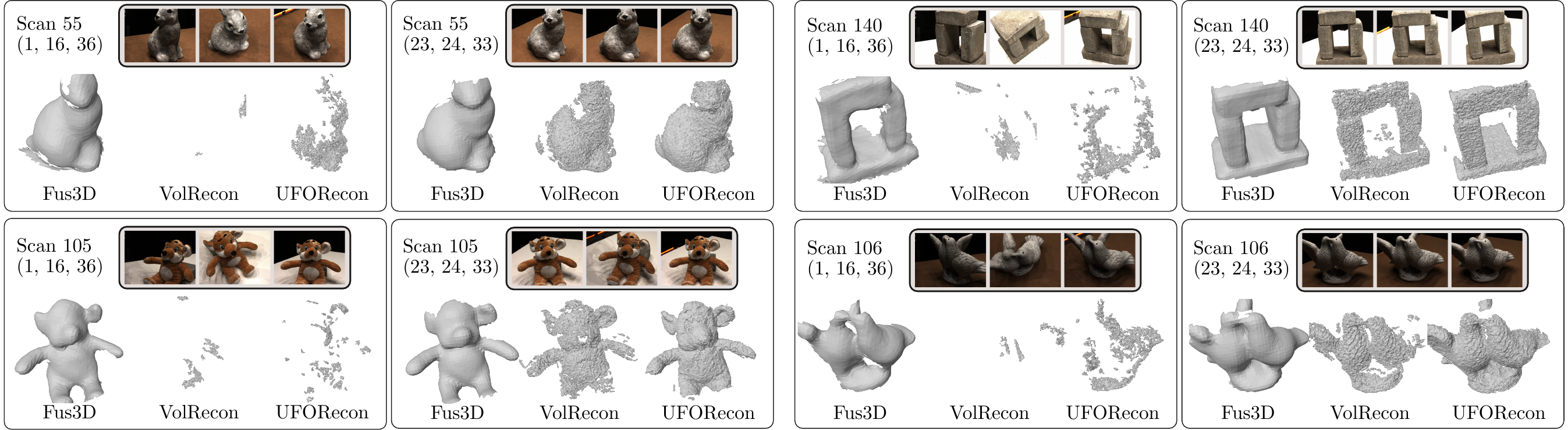}
    \caption{Qualitative results on \textsc{DTU}. Beige boxes indicate input views. Numbers in brackets correspond to image indices of ``favorable'' (23, 24, 33) and ``unfavorable'' (1, 16, 36) view combinations. (Following the baselines' evaluation protocols, only visible geometry is evaluated.)}
    \label{fig:dtu_qual}
\end{figure}

\begin{wraptable}{r}{0.6\textwidth}
    \tiny
\centering
\setlength\tabcolsep{2pt}
\begin{tabular}{l|ccc|ccc}
                    & \multicolumn{3}{c|}{ favorable (23, 24, 33)\quad  } & \multicolumn{3}{c}{ \quad unfavorable (1, 16, 36)  \quad  } \\
                & {$\downarrow$} CD \quad & D$_\text{GT2P}$ & D$_\text{P2GT}$ \quad&\quad CD \quad & D$_\text{GT2P}$ & D$_\text{P2GT}$ \\
\hline \hline
VolRecon $\dagger$& 2.905 & 2.331 & 3.478 & 5.781 & 3.811 & 7.751 \\
UFORecon $\dagger$& 2.771 & 2.101 & 3.440 & 3.275 & 2.252 & 4.298 \\\hline
VolRecon & 3.678 & 3.369 & 3.987 & 8.878 & 5.869 & 11.887 \\
UFORecon & 3.415 & 2.989 & 3.841 & 4.942 & 3.737 & 6.146 \\
Fus3D & 2.432 & 1.804 & 3.059 & 3.525 & 2.814 & 4.236 \\
        
    \end{tabular}
    \caption{Quantitative results with direct SDF prediction baselines in a feed-forward scenario. For reference, $\dagger$ indicated non-feed forward (COLMAP) poses for competing methods.}
    \label{tab:dtu_quant}
\end{wraptable}
As shown in Tab.~\ref{tab:dtu_quant} and Fig.~\ref{fig:dtu_qual} (and in \arxivorelse{App. Tab.~\ref{tab:suppl_dtu}}{the supplemental material}),
we consistently outperform both VolRecon and UFORecon in the feed-forward setting with unknown poses. We improve Chamfer distance by 29\% in both the favorable and unfavorable cases.
Qualitatively, our reconstructions are more complete and less noisy throughout; this is most pronounced in the unfavorable setting.
Notably, despite requiring no camera parameters at inference time, we remain competitive with both baselines even when supplied with COLMAP-estimated poses. 
This suggests that the joint multi-view prior preserved in our volumetric latent space effectively compensates for the absence of explicit camera conditioning.

\subsubsection{Comparison with Unposed Predict-Then-Fuse Baselines\\}
\label{subsec:eval_objaverse}

\hspace{-0.5em}We compare our method with the default 2D extraction paradigm of geometry transformers on the \textsc{Objaverse} dataset, isolating the effect of direct volumetric latent extraction versus predict-then-fuse surface reconstruction. 

For a fair comparison, we fine-tune VGGT on the \textsc{Objaverse} training set (denoted VGGT$_\text{ft}$), supervising depth maps and camera parameters following our
\begin{wrapfigure}{r}{0.55\textwidth}
    \begin{minipage}[c]{0.48\linewidth}
    \includegraphics[width=\textwidth]{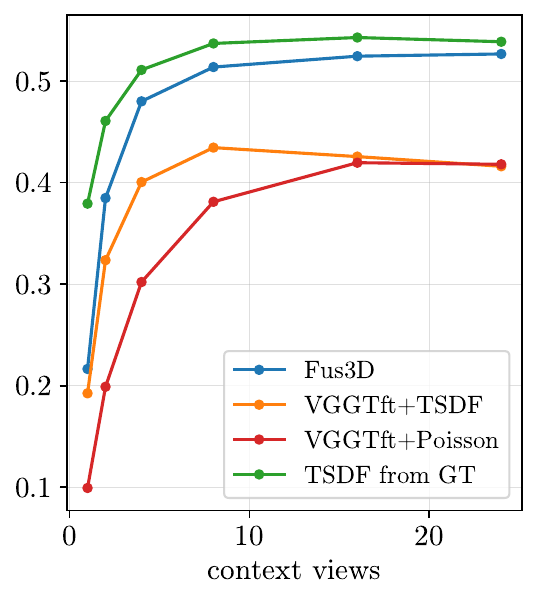}
    \end{minipage}
    \begin{minipage}[c]{0.48\linewidth}
    \includegraphics[width=\textwidth]{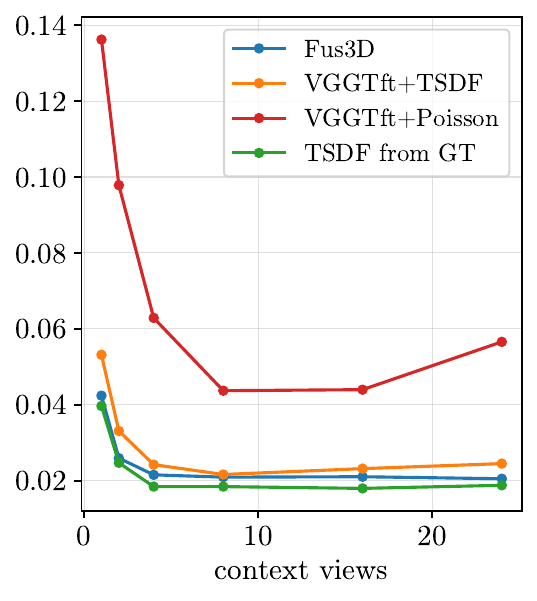}
    \end{minipage}
    \caption{Comparison vs \emph{ground-truth TSDF isosurfaces} (green) on 170 \textsc{Objaverse} test scenes, w.r.t number of input views. Left: F-scores with threshold~$0.5\eps$. Right: Chamfer distances.}
\label{fig:objaverse_quant}
\end{wrapfigure}
standard training regime, with conventional fine-tuning for the prediction heads and LoRA-based backbone refinement~\cite{ren2025fin3r}. 
We compare with two post-hoc fusion strategies for VGGT$_\text{ft}$ outputs: \emph{VGGT$_\text{ft}$+Poisson},  applying Poisson surface reconstruction~\cite{kazhdan2006poisson} to predicted and fused point maps, and \emph{VGGT$_\text{ft}$+TSDF}, constructing isosurfaces from predicted depth maps and cameras via TSDF fusion~\cite{millane2024nvblox} followed by marching cubes~\cite{open3d}.
For Fus3D, surfaces are likewise extracted via marching cubes. 
To ensure a fair comparison, all methods are evaluated at matching output resolutions by sampling with 
the same 
voxel size~$\eps$.
Reconstructed surfaces are compared against ground truth via F-score and Chamfer distance computed on sampled surface points, following Yariv et al.~\cite{yariv2024mosaicsdf}.

\begin{figure}
    \includegraphics[width=\textwidth]{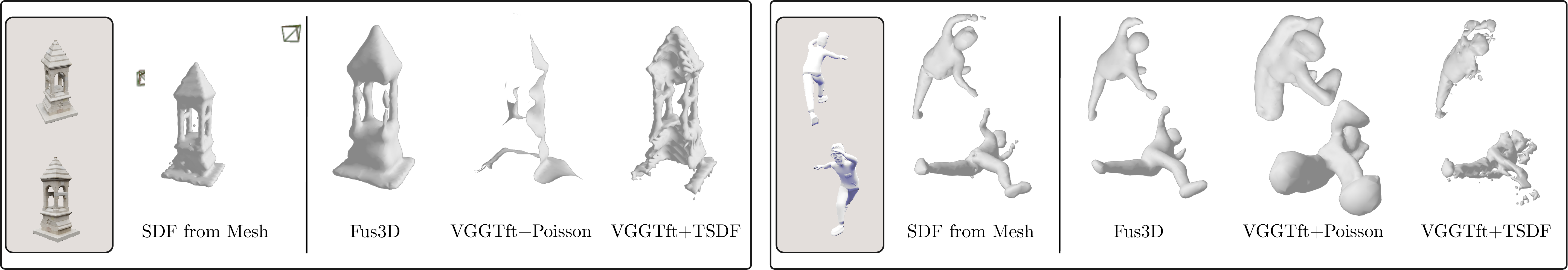}
    \caption{Visual examples from two input views depicted in the beige boxes.}
    \label{fig:eval_objaverse_sparse}
\end{figure}

\begin{figure}
    \includegraphics[width=\textwidth]{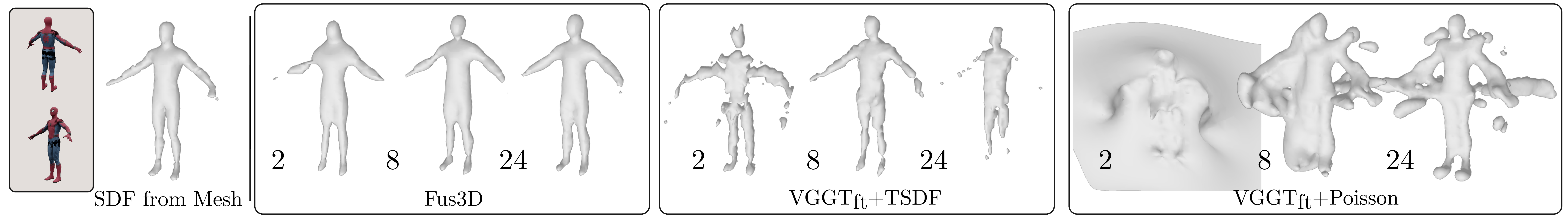}
    \caption{Visual examples with varying number of input views. Two exemplary views shown on the left.}
    \label{fig:eval_objaverse_various}
\end{figure}

Fig.~\ref{fig:objaverse_quant} reports these metrics as a function of the number of input views. 
As \textsc{Objaverse} is a 360\textdegree~object-centric dataset, large regions remain unobserved in sparse-view settings, directly exposing the limitations of predict-then-fuse approaches: TSDF fusion leaves holes in unobserved regions, and while Poisson reconstruction can close smaller gaps, its extrapolation is limited and not robust. 
We include a TSDF constructed from ground-truth depth and cameras, \emph{TSDF from GT}, indicating the upper bound achievable without geometric extrapolation.
Visual examples are shown in Fig.~\ref{fig:eval_objaverse_sparse} and \ref{fig:eval_objaverse_various}. 
Notably, although both Fus3D and VGGT$_\text{ft}$ are trained on up to 8 views, Fus3D generalizes well to 24 views at inference time, while the predict-then-fuse baselines suffer from accumulating inconsistencies at higher view counts, directly validating the scalability advantage of extracting geometry from the joint latent space rather than fusing per-view predictions.

Tab.~\ref{tab:objaverse_metrics} reports full metrics at 8 input views across 170 \textsc{Objaverse} test scenes. 
\begin{wraptable}{r}{0.45\textwidth}
    \begin{minipage}[c]{\linewidth}
    \tiny
    \begin{tabular}{l|ccccc}
                       & F$_\eps$ & F$_{0.5\eps}$ & CD    & EMD   & SDF  \\
                       & & & & & MAE \\
                       \hline
    Fus3D       &\textbf{ 0.83}     & \textbf{0.51}          & \textbf{0.021} & 0.019 & \textbf{0.004}   \\
     VGGT$_{ft}$+
     TSDF   & 0.80     & 0.43          & 0.022 &\textbf{ 0.014} & 0.023   \\
    VGGT$_{ft}$+
    Poiss.  & 0.70     & 0.38          & 0.044 & 0.052 & --      \\
    \hline
    TSDF VGGT          & 0.35     & 0.12          & 0.080 & 0.082 & 0.136   \\
    Poiss. VGGT        & 0.29     & 0.10          & 0.121 & 0.788 & --      \\
    \end{tabular}
    \end{minipage}
    
    \caption{Quantitative comparison on the 170 test scenes of the \textsc{Objaverse} dataset for 8 input views. Lower segment shows non-finetuned baselines.}
    \label{tab:objaverse_metrics}
\end{wraptable}
Fus3D outperforms all baselines on F-score at both thresholds, Chamfer distance, and mean absolute error~(MAE) on valid SDF values of $\grid$, with a particularly pronounced advantage on SDF MAE (0.004 vs. 0.023 for VGGT$_\text{ft}$+TSDF), a direct consequence of regressing the SDF explicitly rather than deriving it from fused depth predictions. 
VGGT$_\text{ft}$+TSDF remains competitive on surface-level metrics but degrades on stricter thresholds (F$_{0.5\eps}$: 0.43 vs. 0.51), reflecting residual noise and incompleteness from per-view fusion.
VGGT$_\text{ft}$+Poisson lags on all metrics, confirming the limited extrapolation capacity of Poisson reconstruction in sparse-view settings.
The substantially lower scores of non-finetuned baselines in the lower segment suggests architectural advantages of Fus3D over VGGT$_\text{ft}$+TSDF and VGGT$_\text{ft}$+Poisson rather than differences in training data or supervision.
The only metric where Fus3D does not lead is EMD, where VGGT$_\text{ft}$+TSDF scores marginally better (0.014 vs. 0.019).

\subsubsection{Ablations and Analysis\\}
\label{subsec:abl_analy}

\begin{figure}
    \centering
    \includegraphics[width=0.7\textwidth]{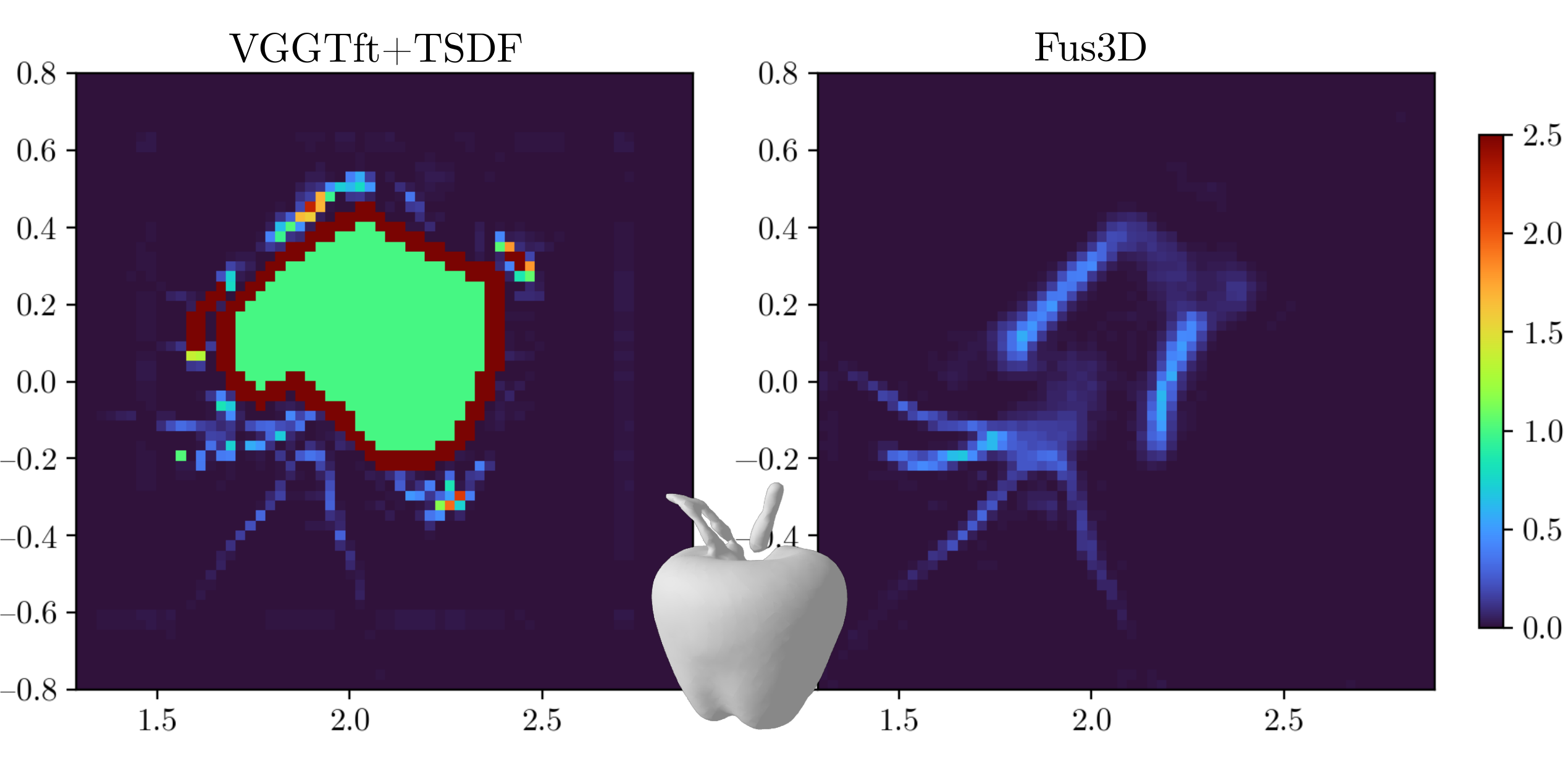}
    \hfill
    \includegraphics[width=0.29\textwidth]{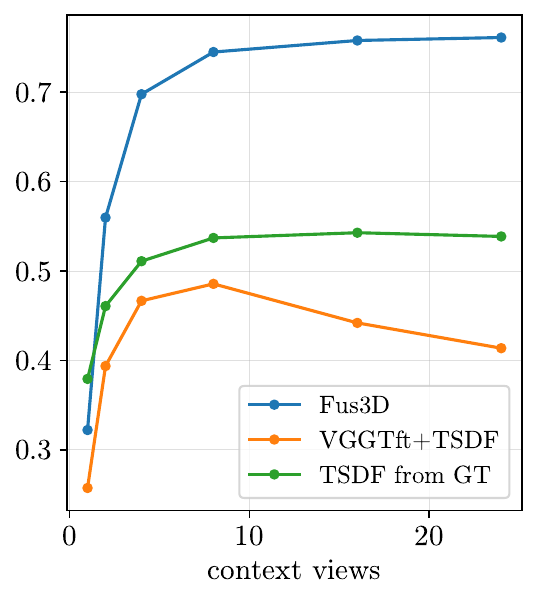}
    \caption{Left: Visualization of the eikonal term for the center slice of the x-axis. Right: Plot of the F$_{0.5\eps}$-score for surfaces extracted at isovalue +0.5$\eps$.}
    \label{fig:eikonal}
\end{figure}

\hspace{-0.6em}\textit{Eikonal Validity.}
Fig.~\ref{fig:eikonal}~(left) visualizes the Eikonal term on a slice of the predicted SDF~$\gsdfpred$ and of the TSDF obtained from VGGT$_\text{ft}$ outputs.
TSDFs store fused projective rather than Euclidean distances, causing systematic distance overestimation and Eikonal violations within the truncation band.
These artifacts manifest as distorted isosurfaces when extracting at offsets $\neq 0$. 
Outside the truncation band, the Eikonal condition is recovered by construction through Euclidian redistancing in observed free space~\cite{millane2024nvblox}. 
Fus3D, by contrast, predicts well-behaved SDF values throughout the volume, particularly near and outside the surface. 
This is reflected quantitatively in Fig.~\ref{fig:eikonal}~(right): at an isovalue offset of $0.5\eps$, Fus3D achieves an even higher F-score than the TSDF constructed from ground-truth synthetic data, demonstrating that the quality of the predicted distance field, not just surface placement, is a meaningful advantage of direct SDF regression.

\begin{figure}[b]
    \includegraphics[width=\textwidth]{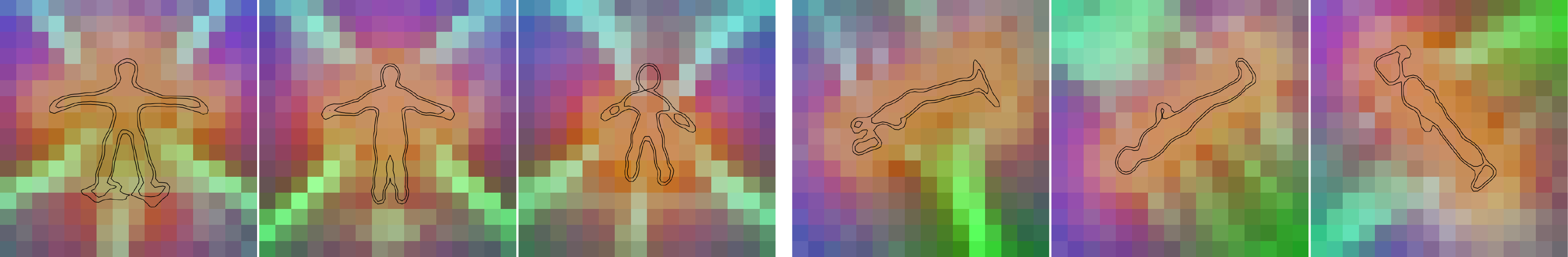}
    \caption{Visualization of the PCA of $\latentthreedpred$ of 3 different avatar-like assets. Images show the central slices, with the first three components used as RGB channels. The contours indicate the zero-crossing of the SDF min-projection along the visualized axis. }
    \label{fig:pca_full}
\end{figure}

\begin{figure}[htb]
    \includegraphics[width=\textwidth]{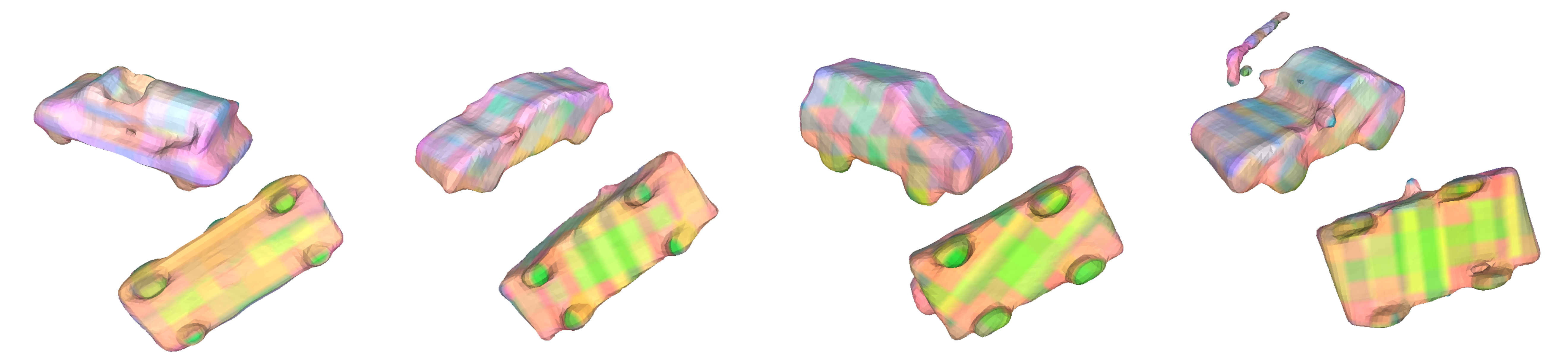}\\
    \hfill
    \caption{PCA Visualization of occupied voxels of $\latentthreedpred$ on four car-like assets. First three components are used as RGB channels and visualized as surface colors. }
    \label{fig:pca_occ}
\end{figure}

\paragraph{PCA Analysis of $\latentthreedpred$.}
To probe the structure of the learned volumetric latent space, we do Principal Component Analysis~(PCA) on $\latentthreedpred$ across objects from the same category.
Fig.~\ref{fig:pca_full} shows that the dominant components vary smoothly in space and correlate with the SDF zero-crossing, adapting consistently to changes in object shape and pose, suggesting that the latent volume encodes geometrically meaningful structure rather than unstructured per-voxel features. 
Fig.~\ref{fig:pca_occ} further shows that features near the surface tend to align with shared geometric parts across instances of the same category, hinting that the latent space captures some degree of category-level surface structure beyond object-specific occupancy. 
We note that due to the relatively low resolution of the volumetric grid, these patterns appear at a coarse spatial scale and finer correspondences are not resolved. 
Nevertheless, the consistent PCA colorings across shapes suggest that the latent volume implicitly organizes geometry in a structured way, which may be beneficial for downstream tasks such as shape recognition, completion or generation.

\paragraph{Inference Evaluation.}

\begin{wrapfigure}{r}{0.33\textwidth}
    \vspace{-0.0cm}
    \includegraphics[width=0.33\textwidth]{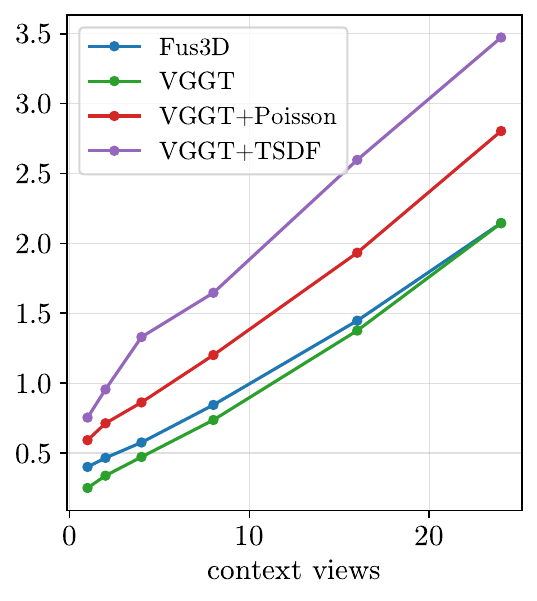}
    \caption{Inference times in seconds w.r.t the number of input views.}
\label{fig:eval_time}
\end{wrapfigure}

Fig.~\ref{fig:eval_time} reports inference times as a function of the number
of input views. 
For VGGT-based baselines, only the relevant heads are evaluated, e.g. $\headd$ and $\headcam$ for TSDF, omitting $\headpts$. 
Both post-hoc fusion approaches add substantial overhead to the total reconstruction time, growing with view count. 
Fus3D introduces a higher constant cost relative to the backbone alone, visible as a larger gap in sparse-view settings; however, this gap narrows as view count increases, since Fus3D omits VGGT's 2D heads $\headd$ and $\headpts$ entirely, leaving the runtime dominated by the shared per-view feature extraction common to all methods. 
At higher view counts, Fus3D therefore scales more favourably than the predict-then-fuse baselines, whose fusion costs grow proportionally with the number of views.

\section{Conclusion}
\label{sec:conclusion}

We present Fus3D, a feed-forward pipeline for dense surface reconstruction that directly regresses Signed Distance Functions from the intermediate feature space of a pretrained multi-view geometry transformer.
To enable training on large-scale real-world data, we introduced a validity-aware SDF supervision scheme that handles non-watertight meshes and unobserved regions without expensive or appearance altering preprocessing. 
By replacing per-view decoder heads with a learned volumetric extraction module, we preserve the joint multi-view prior assembled by the backbone.
By doing so, we address two fundamental failure modes of predict-then-fuse pipelines: sparse-view incompleteness and accumulation of inaccuracies under many views.
Experiments demonstrate consistent improvements in surface completeness, SDF quality, and scaling behavior over both generalizable SDF baselines and feed-forward fusion pipelines.

\section{Acknowledgments}
We would like to thank all members of the Visual Computing Lab Erlangen for the fruitful discussions.

The authors gratefully acknowledge the scientific support and HPC resources provided by the National High Performance Computing Center  of the Friedrich-Alexander-Universität Erlangen-Nürnberg ( NHR@FAU ) under the projects b212dc, b175dc, and b201dc. NHR funding is provided by federal and Bavarian state authorities. NHR@FAU hardware is partially funded by the German Research Foundation (DFG) – 440719683.

Linus Franke was supported in part by the ERC Advanced Grant NERPHYS (101141721, {https://project.inria.fr/nerphys}). The authors are responsible for the content of this publication.

\bibliographystyle{splncs04}
\bibliography{main}

\clearpage
\appendix
\chapter*{Appendix}

\section{Further Details on Experimental Setup} 

\subsection{Datasets}
\label{supp:datasets}

We evaluate our pipeline on two object-centric datasets.
We use the established \textsc{Dtu} dataset to compare against closely related baselines that directly regress SDFs for isosurface reconstruction.
In addition, we use the synthetic \textsc{Objaverse} datset with 360\textdegree~outside-in capture. 
On this dataset, we perform ablations over the number of input views, the supervision signal, and the design of our output SDF, and latent representation presented in the main paper and the supplementary material.

In the following, we provide additional details on the setup and processing of both datasets in addition to the description in the main paper.

\subsubsection{Setup of \textsc{Objaverse}}
Given practical resource constraints, we setup $\approx5\%$ of the \textsc{Objaverse}~\cite{objaverse} subset used by Xiang et al.~\cite{trellis}, comprising 17K object-centric scenes with 24 views each, plus 170 held-out scenes for evaluation. 
Note that this setup requires the photorealistic rendering of all views using Blender as these are not provided.
All objects of this dataset are within $[-0.5, 0.5]^3$ by default. 

For rendering, we again follow the implementation of Xiang et al.~\cite{trellis}.
The views are uniformly distributed on a sphere with radius 2, pointing inwards to the object at the scene center.
We slightly modify the original script by reducing the number of samples per pixel to 32 (with denoising enabled), limiting each scene to 24 views, and exporting depth maps as an additional modality.

Because this is an outside-in capture setup, the network input depends only on externally visible geometry, whereas interior structures remain unobserved but would still strongly affect mesh-derived SDFs.
We therefore remove or ''carve out'' occluded geometry and retain only potentially visible triangles.
To this end, we subdivide triangles to a maximum size of $4\eps$, where $\eps$ is the voxel size our SDF grid, and determine the potentially visible subset by shooting several million rays inward from a bounding sphere.
The resulting carved mesh is then used for SDF supervision during training.

\subsubsection{Setup of \textsc{Dtu}}
Our setup follows VolRecon~\cite{ren2023volreconvolume} and UFORecon~\cite{na2024uforecongeneralizable}, resulting in a split of 119 training scenes and 15 evaluation scenes.
The \textsc{Dtu}~\cite{dtu} dataset provides 49 front-facing views per scene.
As a result, objects are only partially observed, which leads to practical considerations discussed in Supp.~Sec.~\ref{supp:training}.

\subsubsection{Discussion on Datasets and Generalization}
\label{supp:dataset_discussion}
In our setting, a non-finetuned VGGT performs noticeably below a task-adapted model.
We therefore focus our evaluations, mindful of practical compute constraints, on in-domain performance and controlled comparison to relevant baselines, rather than conflating them with inter-dataset generalization.
With sufficiently diverse object-centric training data, the underlying backbone should support substantially stronger cross-dataset transfer.

Concretely, we retain VGGT's convention of defining the scene coordinate frame with respect to the first input view, but unlike the original VGGT setup, we use dataset-level scene scale rather than the mean depth of the first image. 
In addition, both \textsc{Dtu} and \textsc{Objaverse} follow capture configurations that resemble a dome around the object. 
This keeps scene scale effectively fixed and restricts the set of plausible camera parameters.

For extrinsics, the regression problem is therefore reduced from a continuous 6-DoF space to a finite set of valid camera arrangements. 
For intrinsics, the model can likewise specialize to a single parameter set in \textsc{Objaverse} or a small family of calibrated settings in \textsc{Dtu}. 
In that sense, our training distribution is narrower than the one targeted by large-scale general-purpose geometry transformers.

\subsection{Training}
\label{supp:training}

\subsubsection{SDF Supervision} 
\label{supp:sdf_supervision}
We supervise with a direct 3D loss on target SDFs. 
Even when ground-truth meshes are available, computing SDFs involves practical considerations that depend on the chosen SDF calculation method.

\begin{figure}
    \includegraphics[width=\textwidth]{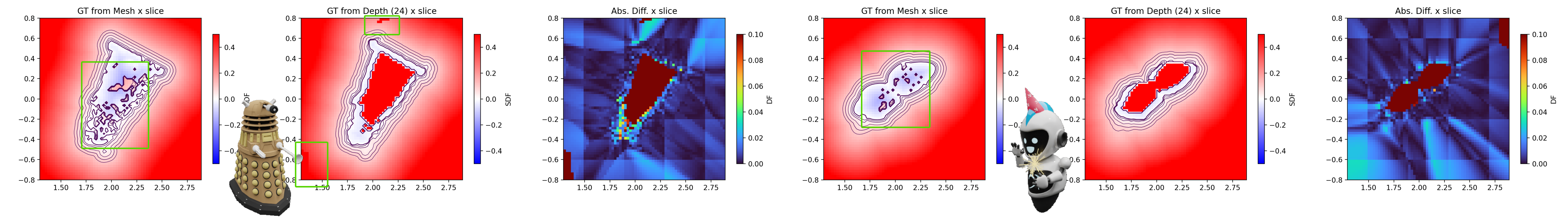}
    \caption{Reliability of SDF from Meshes vs TSDFs from Depth. Non-watertight meshes may lead to sign flips, while TSDFs contain unobserved regions (green boxes). Difference images visualize differences also in values between the two.}
    \label{fig:masking}
\end{figure}

\paragraph{SDF from meshes}
The unsigned distance at a 3D query position is computed as the minimum distance to the mesh surface, obtained by finding the closest point on all triangles.
In a second step, the sign is determined by shooting one or more rays and testing whether the number of mesh intersections is even (outside) or odd (inside).
In theory, a single ray is sufficient to determine the sign, but non-watertight meshes can lead to inconsistent results.
For example, ground planes, self-intersections, or holes make the distinction between interior and exterior ill-defined.
Since most assets in large-scale datasets are not watertight, the sign is often estimated by majority voting across multiple rays.
This can still lead to spurious sign flips; see Supp. Fig.~\ref{fig:masking}.
These sign flips also become visible when extracting isosurfaces from the resulting SDFs; see Supp. Fig.~\ref{fig:grid_signflips}.

\paragraph{Truncated SDFs from rendered depth maps}
\begin{wrapfigure}{r}{0.35\textwidth}
    \begin{minipage}[c]{\linewidth}
        \includegraphics[width=0.48\textwidth]{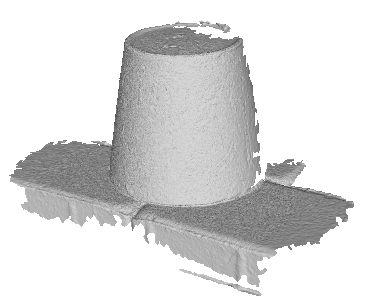}
        \includegraphics[width=0.48\textwidth]{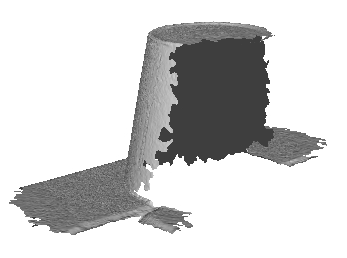}
        
    \end{minipage}
    \caption{Visual example of \textsc{Dtu} ground-truth mesh.}
    \label{fig:dtu_supervision}
\end{wrapfigure}
In more extreme cases, such as the mesh annotations of \textsc{Dtu}, surfaces are open and SDF signs cannot be determined reliably from the mesh alone.
In such cases, a practical alternative is to fuse depth maps and camera parameters into a truncated signed distance field (TSDF).
This approach is also more broadly applicable beyond synthetic data, since real-world datasets provide depth observations more frequently than clean meshes.
We use the state-of-the-art nvblox implementation~\cite{millane2024nvblox}, adapted to run directly in our training environment without precomputation.
To preserve details, we fuse at higher resolution and then sample the TSDF at the target resolution, both for supervision and for baseline comparisons.

For supervision, we fuse all available views of a scene to allow extrapolation beyond the visible regions of the input views, while  still-unobserved voxels are marked as invalid.

However, TSDF fusion stores projective rather than true Euclidean distances.
As a result, the zero-level set is represented correctly, but voxel values generally overestimate the minimal distance to the surface and therefore violate the Eikonal condition.
This limits e.g. surface extraction at arbitrary isovalues and can affect downstream tasks that rely on accurate SDF values.

\paragraph{Validity masking}

We address these practical issues during SDF supervision by targeting robustness of loss computations.
See Supp. Fig.~\ref{fig:masking} for visualizations.

In related experiments (see Supp.~Sec.~\ref{supp:misc_results}), we found that supervision with TSDFs alone degrades performance.
For \textsc{Dtu}, we therefore use a hybrid strategy:
we compute $\gL_\textrm{SDF}(\grid)$ from a depth-fused TSDF to obtain reliable signs, while retaining mesh-based supervision for $\gL_\textrm{SDF}(\nearsurface)$ to preserve finer detail.
For the latter, we use the unsigned variant, which is implemented by setting $\maske$ to the empty set during computation.

More generally, sign flips in mesh-derived SDFs often induce violations of the Eikonal condition.
As described in Main Sec.~\ref{sec:training}, we handle these regions by masking out voxels that violate the Eikonal constraint when computing $\gL_\textrm{SDF}$ and falling back to the unsigned variant instead.
Gradient regression losses are likewise unreliable in these regions and are therefore disabled.
Specifically, $\maske$ is constructed by evaluating the Eikonal term on the target SDF grid and propagating Eikonal violations to neighboring voxels by applying a 3D max-pooling operation with kernel size 5, followed by thresholding with 2.

\subsubsection{Alignment of Reference Frames}
\label{supp:alignment}
When comparing surfaces and point clouds from methods that predict in different reference frames and scales, a 7-DoF similarity transform must be estimated to enable a fair comparison.
We follow the strategy of Knapitsch et al.~\cite{knapitsch2017tankstemples} and estimate this transform by registering predicted camera poses to the ground-truth poses.
We found this to be more robust than non-rigid ICP~\cite{besl1992icp} combined with RANSAC, especially when scene parts are severely misaligned (see the non-fine-tuned VGGT examples in Supp.~Fig.~\ref{fig:grid_various}).
For the \emph{On-} and \emph{Offline} scenarios in our \textsc{Dtu} experiments, we additionally apply rigid ICP to further improve alignment for all methods.

\subsubsection{Details on Metrics}
\label{supp:metrics}
For the metrics reported in Main Tab.~\ref{tab:objaverse_metrics} and Main Figs.~\ref{fig:objaverse_quant} and~\ref{fig:eikonal},
we use the normalized Earth Mover's Distance also used by Zhou et al.~\cite{zhou2021shapegen}, based on Mo's implementation~\cite{pytorchEmd}.
For these results, we use Taheri's implementation~\cite{pytorchChamferD} for the Chamfer distance.
SDF MAE is computed only on valid samples.

For the metrics reported in Main Tab.~\ref{tab:dtu_quant} and Supp. Tab.~\ref{tab:suppl_dtu}, we reuse the evaluation script provided by VolRecon~\cite{ren2023volreconvolume} and UFORecon~\cite{na2024uforecongeneralizable}, whose Chamfer distance implementation is based on Scikit-learn~\cite{pedregosa2011scikit}.

\section{Extended Results}

\subsection{Comparison to Feed-Forward SDF baselines on \textsc{Dtu}}
\label{supp:eval_dtu}

\begin{table}
    \centering
    \tiny
\begin{tabular}{l|ccccccccccccccc|r}
Method & 24 & 37 & 40 & 55 & 63 & 65 & 69 & 83 & 97 & 105 & 106 & 110 & 114 & 118 & 122 & Mean \\
\hline\hline
\emph{Favorable} &&&&&&&&&&&&&&&& \\
\hline
\emph{GT} &&&&&&&&&&&&&&&& \\
 VolRecon & 1.652 & 2.612 & 1.704 & 1.190 & 1.538 & 2.027 & 1.340 & 1.532 & 1.412 & 1.046 & 1.383 & 1.739 & 0.883 & 1.324 & 1.322 & 1.514 \\
 UFORecon & 0.774 & 2.105 & 1.348 & 0.858 & 1.154 & 1.160 & 0.709 & 1.243 & 1.172 & 0.808 & 0.890 & 0.568 & 0.521 & 0.869 & 0.980 & 1.011 \\
\emph{Offline} &&&&&&&&&&&&&&&& \\
VolRecon & 2.284 & 3.244 & 2.484 & 2.537 & 4.200 & 4.108 & 2.645 & 3.580 & 2.817 & 2.446 & 2.449 & 3.088 & 2.238 & 2.546 & 2.903 & 2.905 \\
 UFORecon & 1.862 & 4.532 & 2.687 & 2.296 & 3.951 & 3.452 & 2.455 & 4.145 & 2.242 & 3.163 & 1.730 & 2.606 & 1.654 & 2.159 & 2.623 & 2.771 \\
\emph{Online} &&&&&&&&&&&&&&&& \\
VolRecon & 3.234 & 4.524 & 3.655 & 3.988 & 4.159 & 4.719 & 3.093 & 3.581 & 3.408 & 3.591 & 3.666 & 3.458 & 3.064 & 3.186 & 3.848 & 3.678 \\
UFORecon & 3.190 & 5.451 & 4.040 & 3.497 & 4.124 & 3.416 & 2.991 & 3.075 & 2.999 & 2.786 & 2.954 & 3.437 & 2.716 & 3.147 & 3.404 & 3.415 \\
Fus3D & 2.636 & 3.129 & 2.606 & 2.131 & 2.432 & 2.267 & 2.506 & 2.833 & 2.464 & 1.977 & 2.079 & 2.057 & 2.026 & 3.092 & 2.239 & 2.432 \\
\hline
\emph{Unfavorable} &&&&&&&&&&&&&&&& \\
\hline
\emph{GT} &&&&&&&&&&&&&&&& \\
VolRecon & 3.846 & 3.745 & 4.788 & 3.691 & 3.261 & 3.963 & 3.053 & 5.021 & 3.822 & 3.076 & 4.171 & 3.475 & 1.952 & 3.824 & 3.875 & 3.704 \\
UFORecon & 1.298 & 1.924 & 1.344 & 1.298 & 1.204 & 1.523 & 1.013 & 1.473 & 1.311 & 1.048 & 1.433 & 0.865 & 0.572 & 1.288 & 1.108 & 1.247 \\
\emph{Offline} &&&&&&&&&&&&&&&& \\
VolRecon & 6.107 & 4.261 & 5.061 & 5.933 & 5.436 & 7.371 & 5.929 & 8.503 & 5.589 & 4.937 & 5.294 & 5.589 & 3.134 & 6.143 & 7.432 & 5.781 \\
UFORecon & 2.860 & 4.164 & 2.665 & 2.802 & 3.113 & 3.837 & 2.359 & 4.316 & 7.326 & 2.477 & 2.740 & 2.497 & 2.089 & 3.072 & 2.813 & 3.275 \\
\emph{Online} &&&&&&&&&&&&&&&& \\
VolRecon & 9.152 & 10.665 & 6.289 & 9.582 & - & 10.952 & 14.509 & 10.022 & 8.944 & 6.915 & 10.013 & 5.681 & 6.741 & 7.252 & 7.578 & 8.878 \\
UFORecon & 4.691 & 6.729 & 5.493 & 4.462 & 10.300 & 3.644 & 3.908 & 5.478 & 3.848 & 5.945 & 4.680 & 3.501 & 3.044 & 4.777 & 3.625 & 4.942 \\
Fus3D & 4.074 & 3.522 & 2.899 & 3.156 & 4.247 & 3.579 & 3.177 & 3.960 & 3.773 & 2.340 & 3.185 & 4.432 & 3.123 & 4.083 & 3.328 & 3.525 \\
\end{tabular}
    \caption{Quantitative results in offline vs online scenarios, and favorable vs unfavorable view combinations, separated by pose-estimation type. Values are Chamfer distances ($\downarrow$\,CD). GT indicates \textsc{Dtu} dataset poses, \emph{Offline} refers to COLMAP estimation, and \emph{Online} to feed-forward methods using VGGT or ours.}
    \label{tab:suppl_dtu}
\end{table}

Here, we provide additional details on the comparison to feed-forward SDF prediction baselines (VolRecon~\cite{ren2023volreconvolume} and UFORecon~\cite{na2024uforecongeneralizable}), as described in Main Sec.~\ref{subsec:eval_dtu}.

\paragraph{Details on Baselines}

Camera calibration for the \emph{Offline} case is performed using COLMAP~\cite{schonberger2016pixelwiseview} with default parameters for feature extraction, exhaustive matching, and mapping over all 49 views.
We also tested running COLMAP using only the three input views, but the mapping step frequently failed due to an insufficient number of matches.

For the baseline methods, we used the best-performing publicly available checkpoints. 
For alignment of all outputs to the ground-truth mesh, see Supp. Sec.~\ref{supp:alignment}.

\paragraph{Further Evaluation}

\begin{wraptable}{r}{0.4\textwidth}
    \centering
    \small
    
\begin{tabular}{l|cc|r}
                & Cam. & SDF ~&~ Total  \\
                \hline\hline
\emph{Offline} &&&\\
UFORecon        & {66.4}    & 31.5 & 97.9 \\
VolRecon        &   {66.4}                        & 21.6 & 88.0 \\
\hline
\emph{Online} &&&\\
UFORecon        & {0.31}   & 31.5 & 31.8 \\
VolRecon        &   {0.31}                        & 21.6 & 21.9 \\
Fus3D    &     --                    & 0.53 & 0.53 \\

\end{tabular}
    \caption{Measured execution time in seconds. "Cam." indicates time for camera parameter estimation required for baselines, SDF for direct SDF estimation.}
    \label{tab:suppl_dtu_times}
\end{wraptable}

In Tab.~\ref{tab:suppl_dtu}, we present metrics of Main Tab.~\ref{tab:dtu_quant} on a per-scene level.
Measured execution time of our and baseline methods are in Supp. Tab.~\ref{tab:suppl_dtu_times}. 
Note that the reported baseline runtimes use a higher input resolution ({\small $512 \times 640$}) than Fus3D ({\small $364 \times 448$}) and are measured with default, quality-oriented settings; they could likely be tuned further for online deployment.
Nevertheless on these settings, our method is about 40$\times$ faster in estimation compared to the baselines.

\subsection{Comparison with Unposed Predict-Then-Fuse Baselines}
\label{supp:eval_objaverse}

\paragraph{Details on Baselines}

The most relevant baseline is VGGT, as our method keeps the same geometry backbone and primarily replaces its 2D decoding and post-hoc fusion with a 3D extraction transformer.

Mindful of dataset size and compute constraints, we adopt a lightweight fine-tuning strategy to setup our baseline \emph{VGGT$_{ft}$} and tune our employed backbone.
Inspired by Ren et al.~\cite{ren2025fin3r}, we update the image encoder (and output heads for VGGT$_{ft}$) while leaving the multi-view aggregation stack unchanged.
Since we apply the exact same strategy to our backbone, the intermediate latent features remain close to those of standard VGGT.
This indicates that the 3D information, which e.g. allowed scene completion, were already represented by the original latent space and only had to be extracted differently.
In order to build up an explicit, scene-consistent 3D state already during image matching, akin to a spatial memory, a more tightly coupled aggregation and extraction is interesting for future work.

\begin{figure}[htb]
    \includegraphics[width=\textwidth]{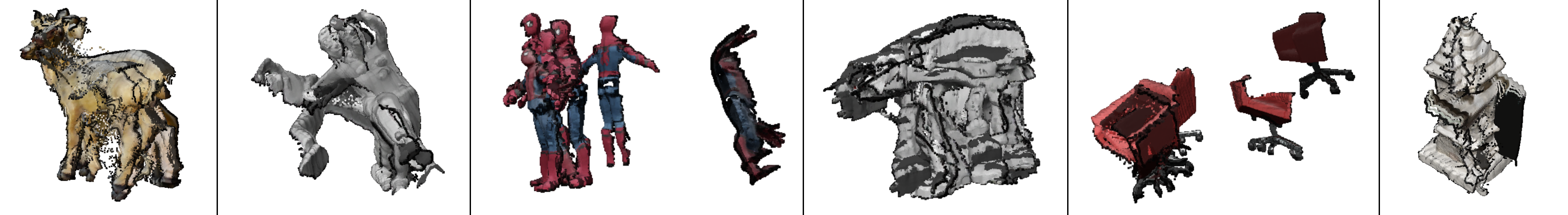}
    \caption{
        Qualitative results of CUT3R on our \textsc{Objaverse} test set.
    \label{fig:cut3r}}
\end{figure}

In the same spirit, we also considered adding CUT3R~\cite{wang2025cut3r} as a representative method with an explicit spatial memory for iterative reconstruction to our main evaluation.
In our setting using non-sequential images, however, its predictions frequently exhibited severe misalignment, which hindered reliable surface reconstruction and coordinate-frame alignment.
We therefore restrict this comparison to a qualitative evaluation in Fig.~\ref{fig:cut3r}.
We hypothesize that CUT3R's sequential processing is a poor fit for our unordered input-view setting, and that its memory mechanism does not compensate for this mismatch.

\paragraph{More visual results}

\begin{figure}
    \includegraphics[width=0.19\textwidth]{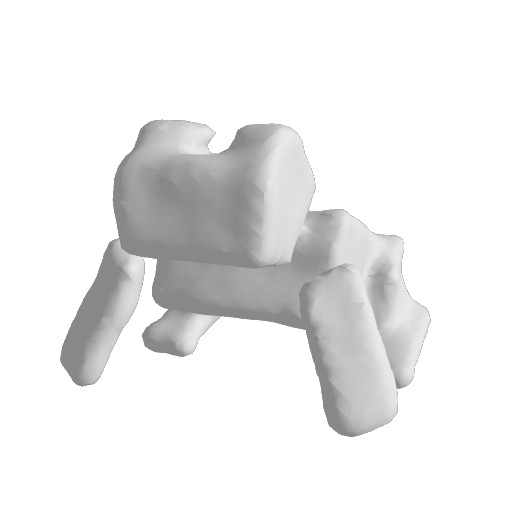}
    \includegraphics[width=0.36\textwidth]{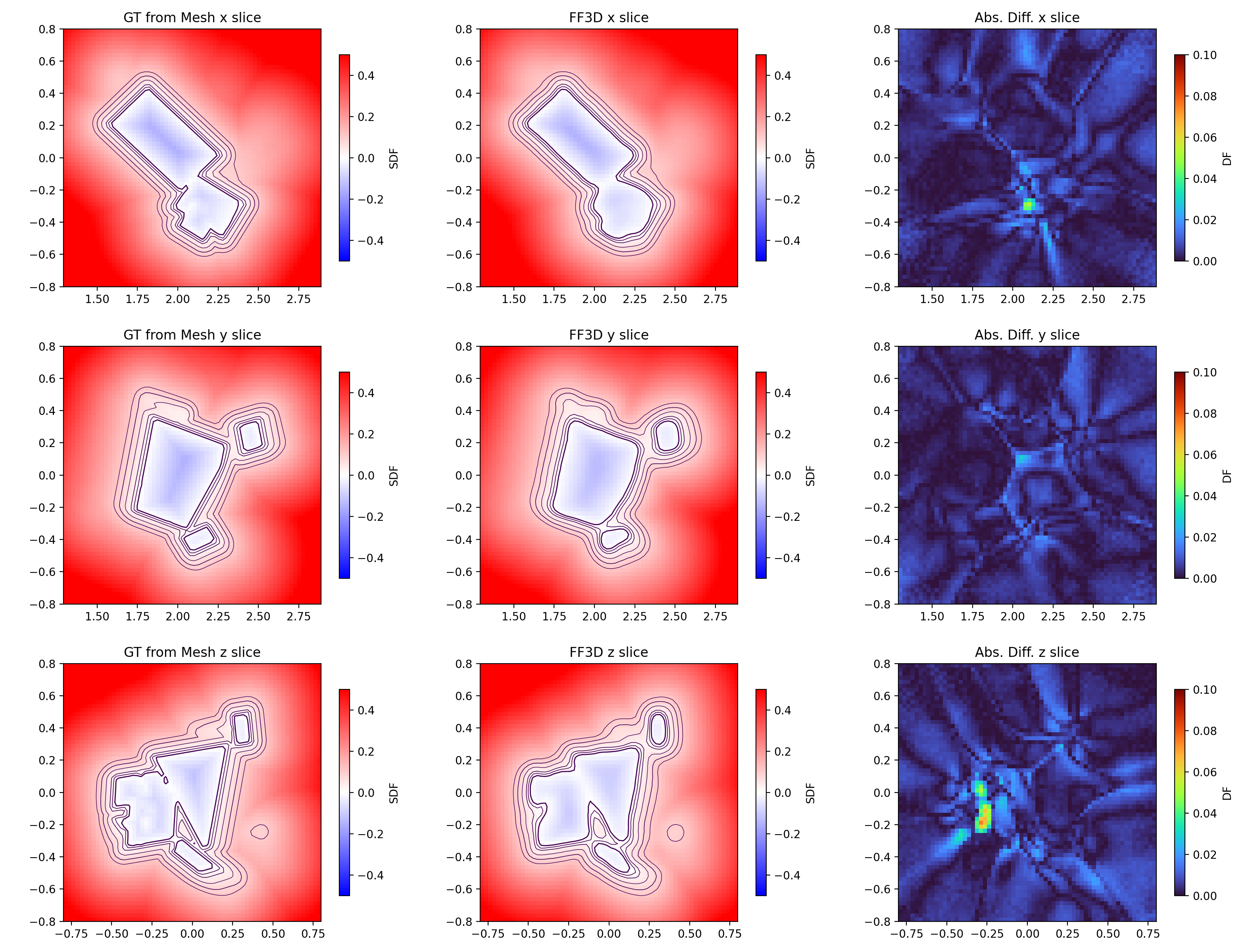}
    \hfill
    \includegraphics[width=0.36\textwidth]{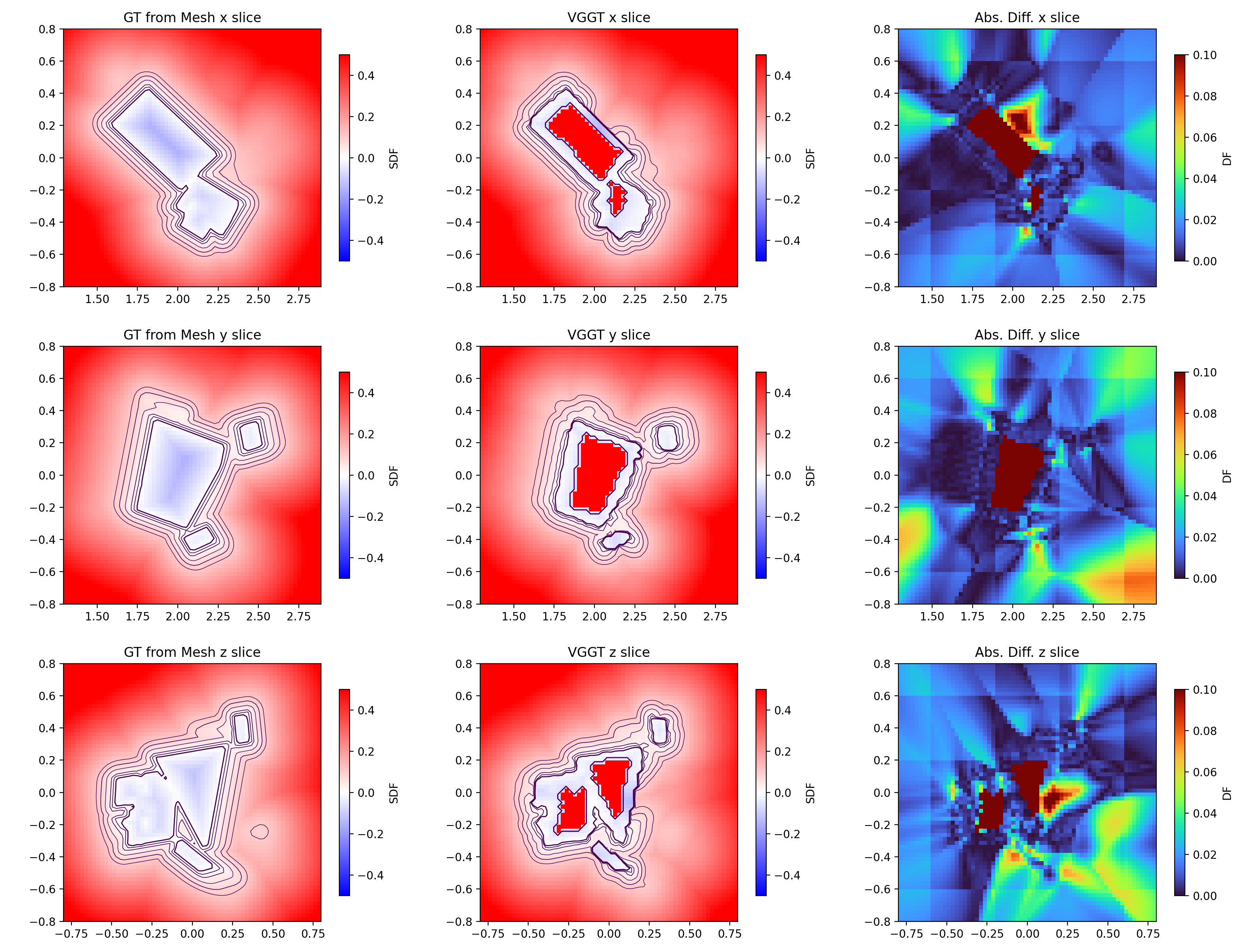}
    \caption{Left: isosurface extracted from our prediction. Center: slices of our predicted SDF and the corresponding error map. Right: slices of the TSDF reconstructed from VGGT$_\text{ft}$ predictions and the corresponding error map.
    \label{fig:slices}}
\end{figure}

\begin{figure}    
    \includegraphics[width=\textwidth]{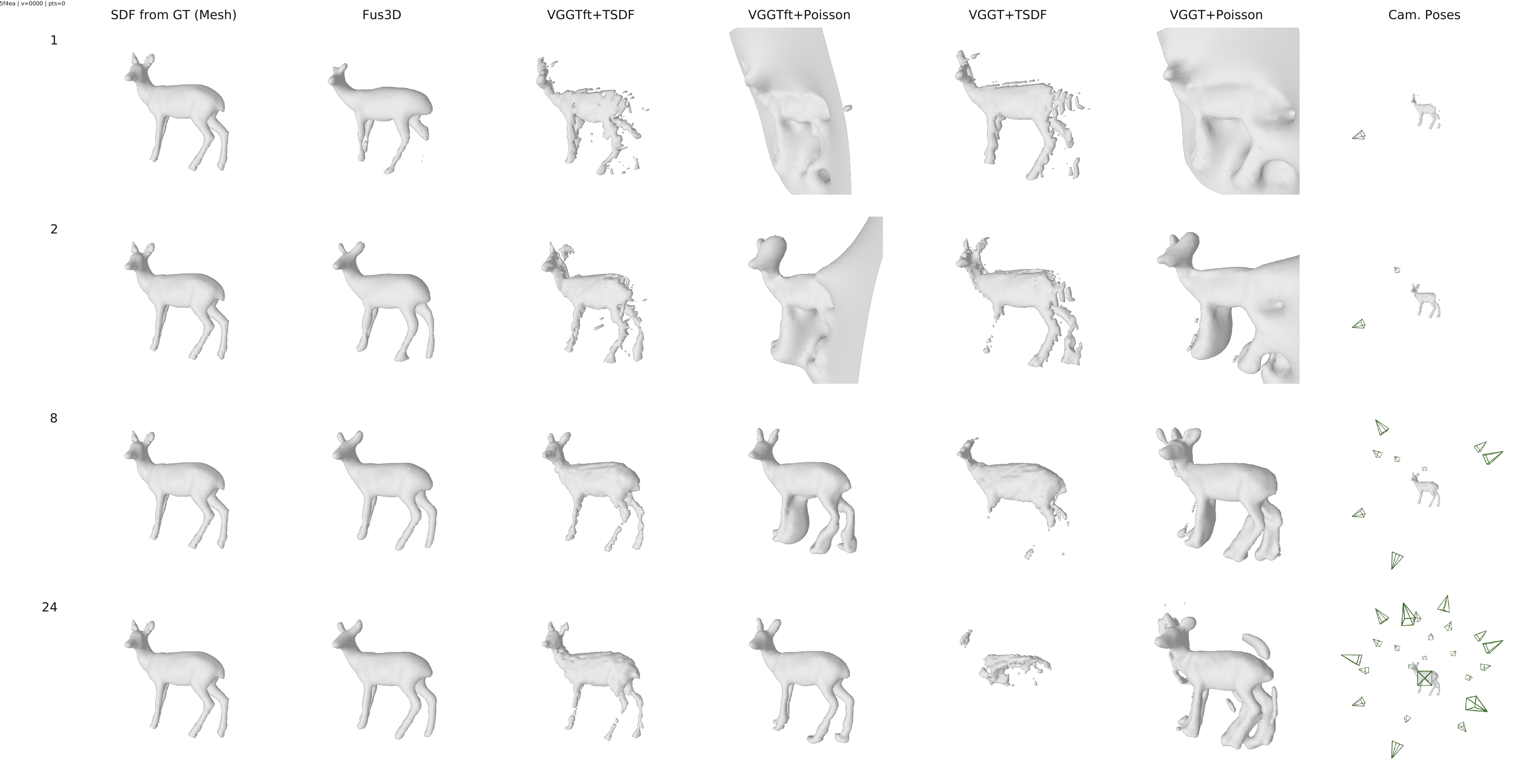}
    
    \caption{Further qualitative results on \textsc{Objaverse}, supplementing Main Fig.~\ref{fig:teaser}.}
    \label{fig:grid1}
\end{figure}

\begin{figure}
    \includegraphics[width=\textwidth]{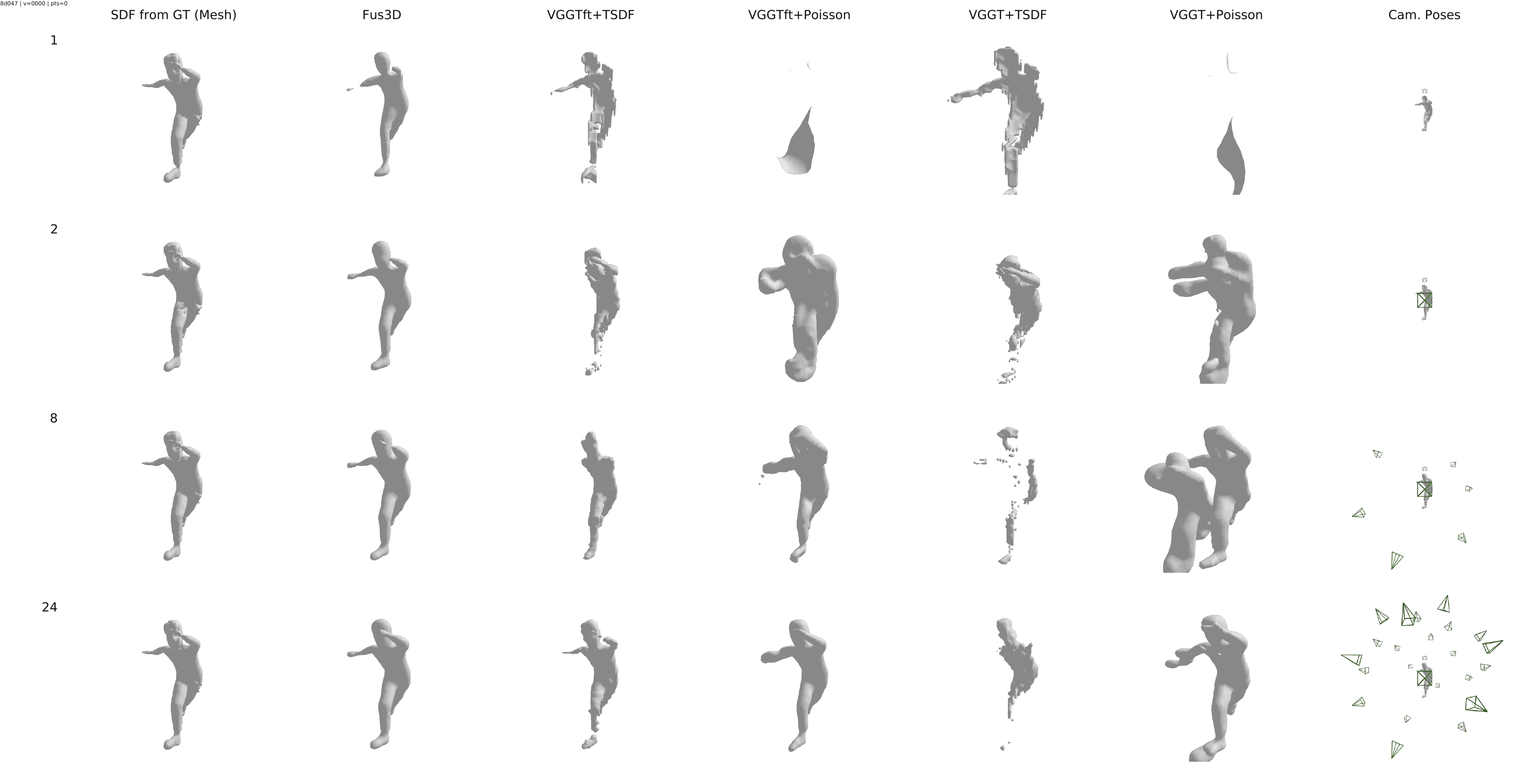}
    
    \includegraphics[width=\textwidth]{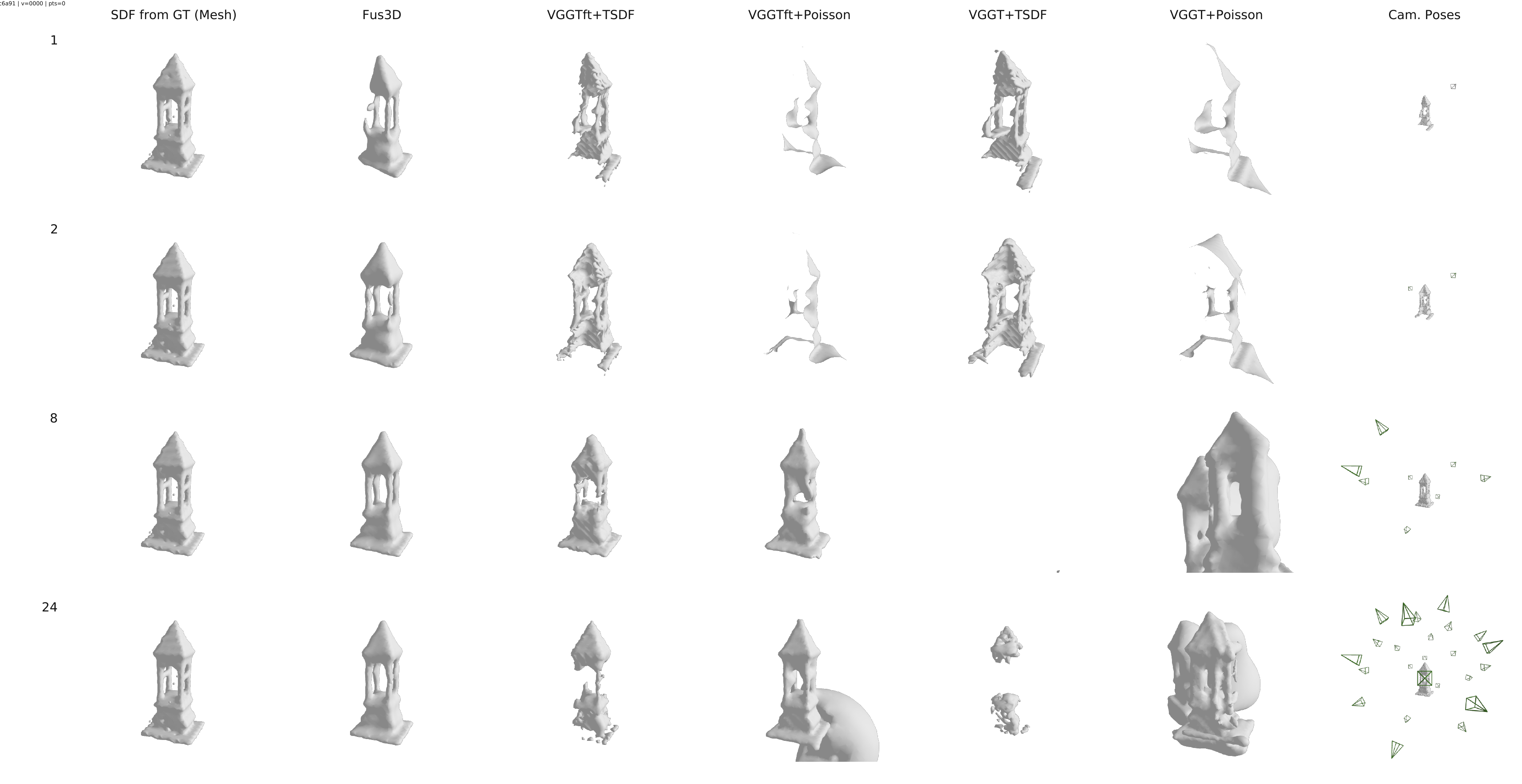}
    \caption{
        Additional qualitative results on \textsc{Objaverse}, supplementing Main Fig.~\ref{fig:eval_objaverse_sparse}.
    \label{fig:grid_various}}
\end{figure}

\begin{figure}
    \includegraphics[width=\textwidth]{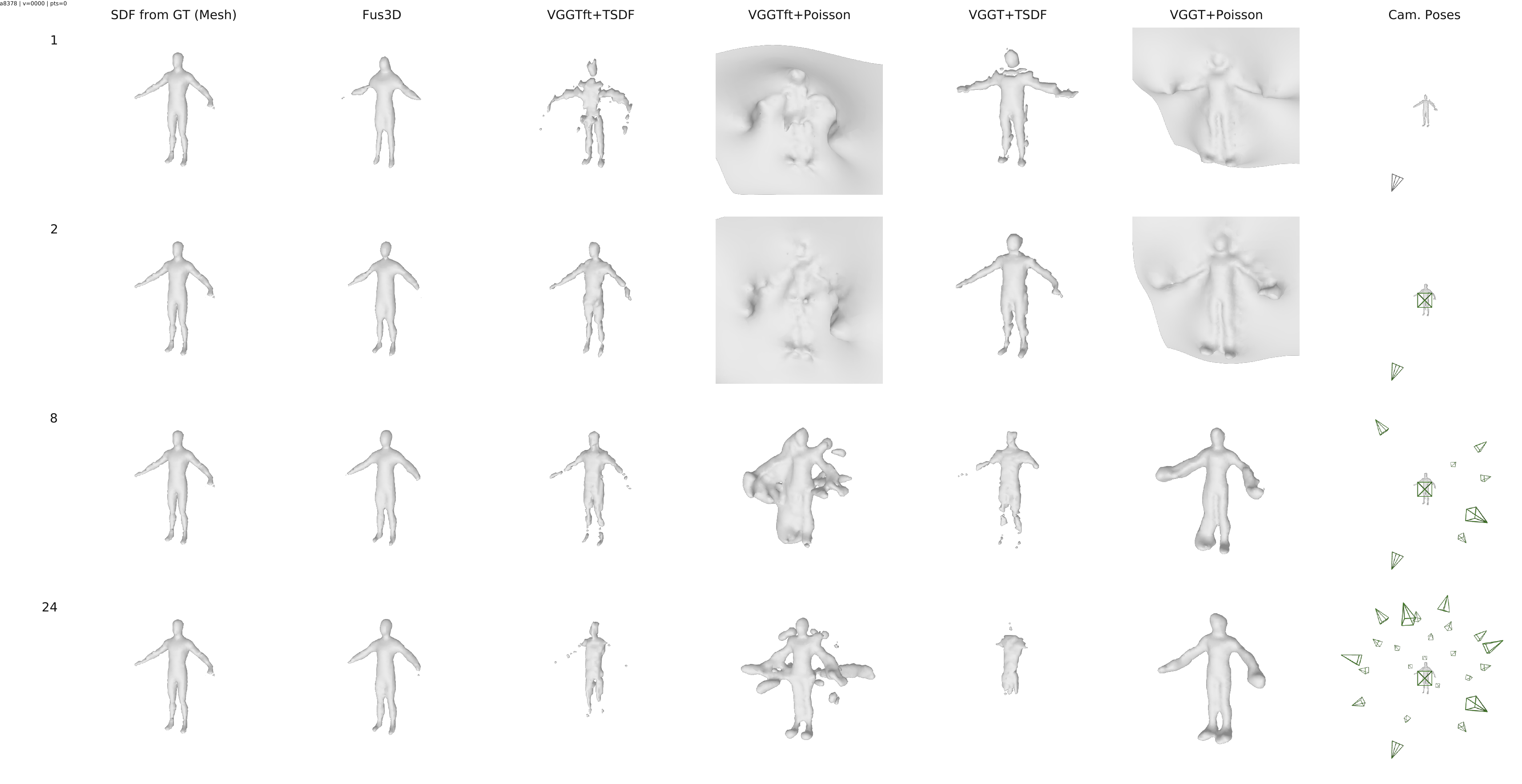}
    
    \includegraphics[width=\textwidth]{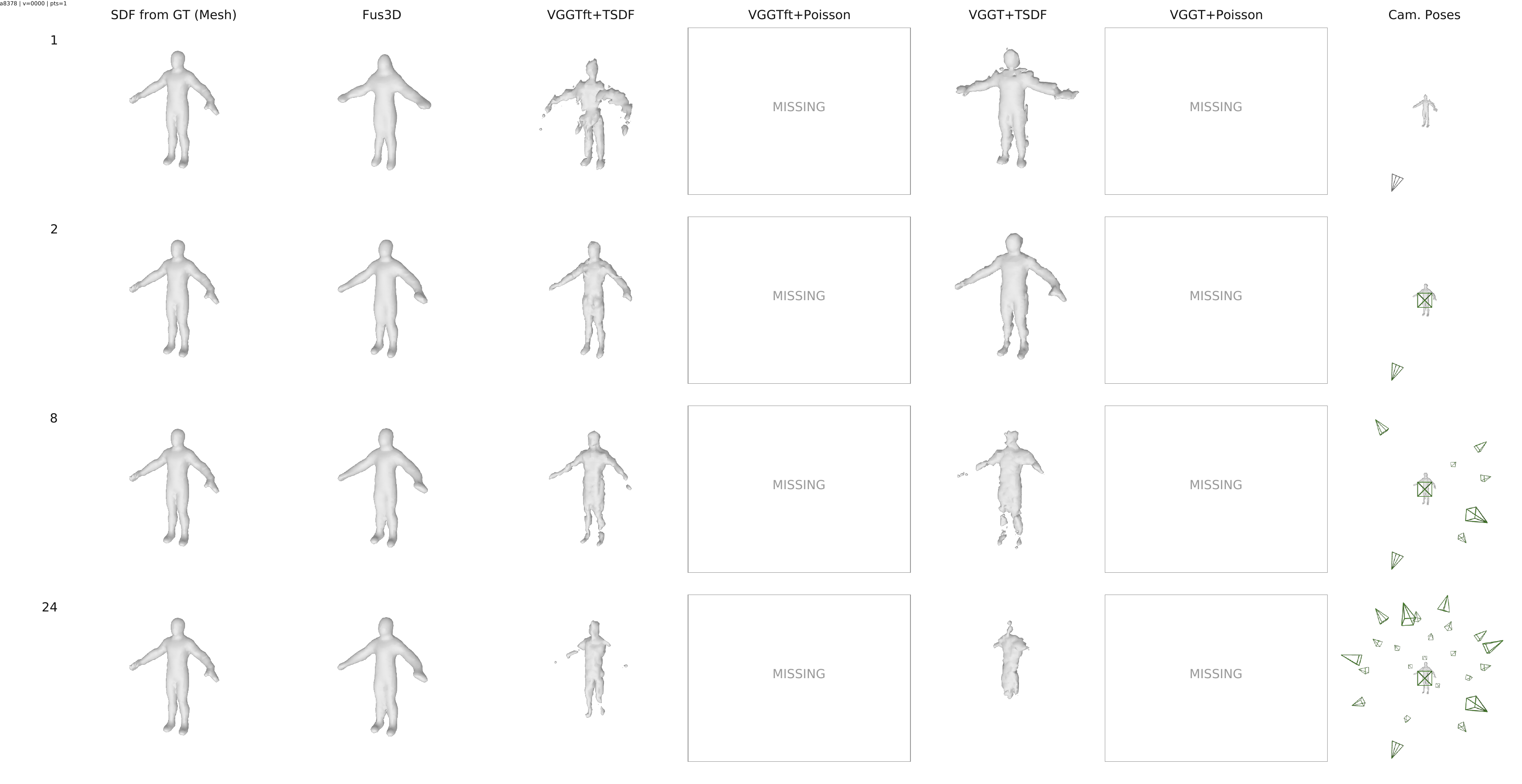}
    \caption{Additional variants of the scene shown in Main Fig.~\ref{fig:eval_objaverse_various}. Top: surface extracted at isovalue $0$. Bottom: surface extracted at isovalue $+0.5\eps$ (Poisson reconstruction is omitted here, as it only reconstructs the zero level set).}
    \label{fig:grid_isoval}
\end{figure}

\begin{figure}
    \includegraphics[width=\textwidth]{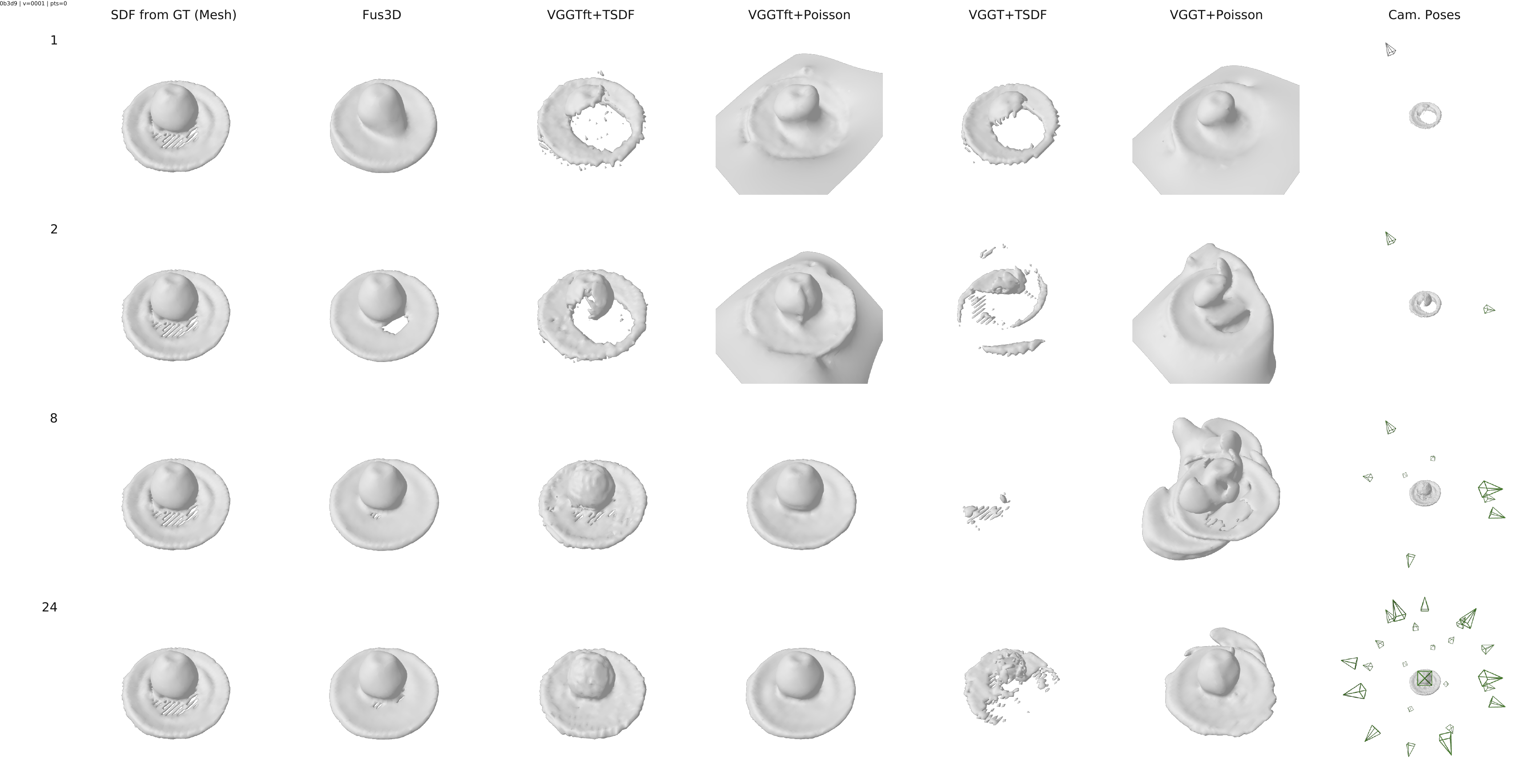}
    \caption{Additional qualitative results on \textsc{Objaverse}.}
    \label{fig:grid2}
\end{figure}

\begin{figure}
    \includegraphics[width=\textwidth]{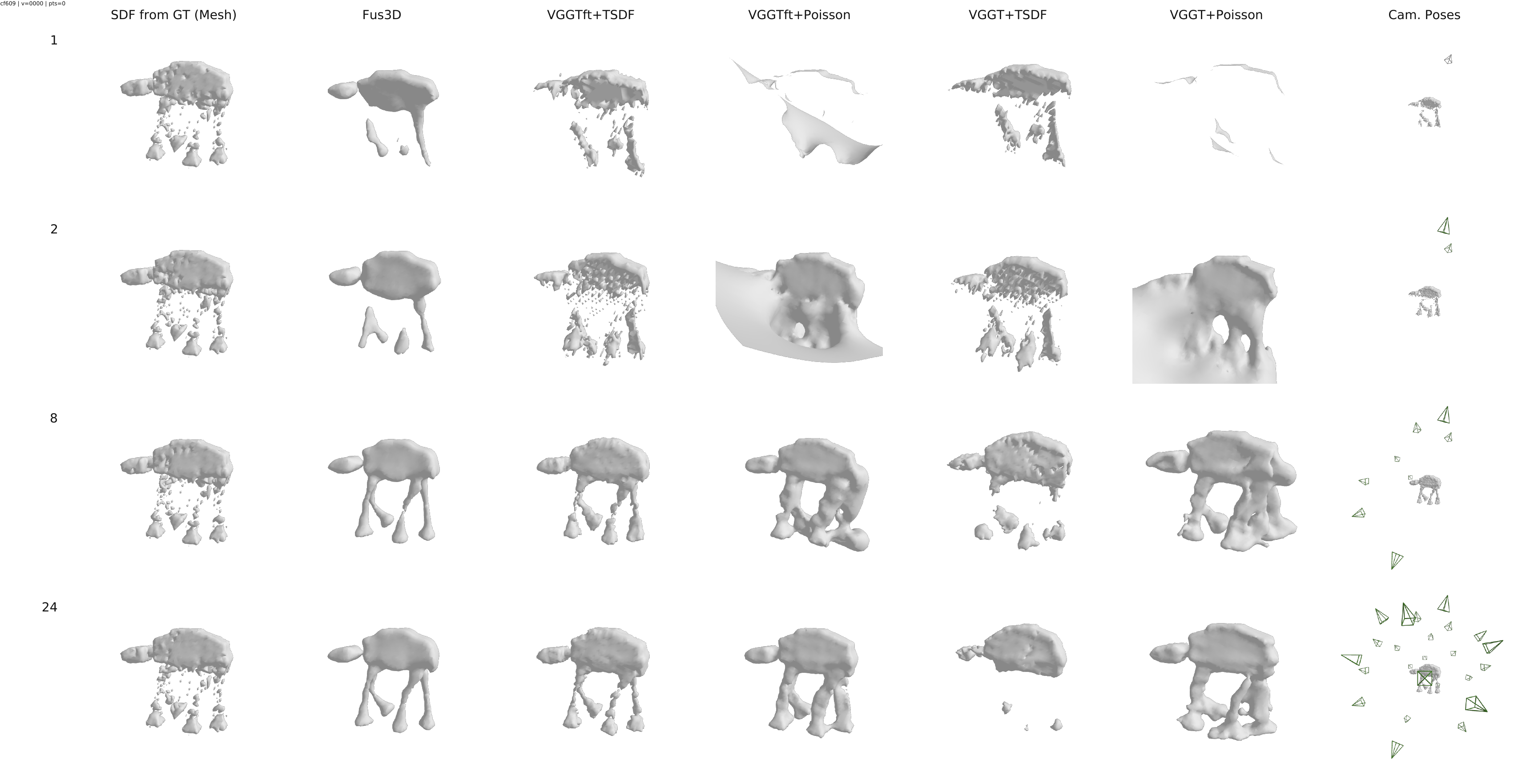}
    \caption{Further qualitative results. Spurious sign flips in mesh-derived SDFs can lead to broken isosurfaces (see, e.g., the legs).}
\label{fig:grid_signflips}
\end{figure}

In Supp. Fig.~\ref{fig:slices}, we visualize slices of the volumetric SDF grids.
Supp. Fig.~\ref{fig:grid_various}, \ref{fig:grid1}, \ref{fig:grid_isoval}, \ref{fig:grid2}, and \ref{fig:grid_signflips} present supplementary qualitative results.

\subsection{Miscellaneous}
\label{supp:misc_results}

\paragraph{3D Querying}

\begin{wraptable}{r}{0.62\textwidth}
    \centering
    \small
    
\begin{tabular}{l|cc|cc}
                & \multicolumn{2}{c}{Variable $\V$} & \multicolumn{2}{c}{Default $\V$} \\
                \hline
                & F$_{0.5\eps}$ & CD & F$_{0.5\eps}$ & CD \\
Fus3D+   & \textbf{0.281} & \textbf{0.018} & \textbf{0.467} & \textbf{0.021} \\
Fus3D+ wo. Q  & 0.274 & 0.021 & 0.466 & 0.022 \\    
\end{tabular}
\end{wraptable}

\begin{figure}
    \includegraphics[width=0.65\textwidth]{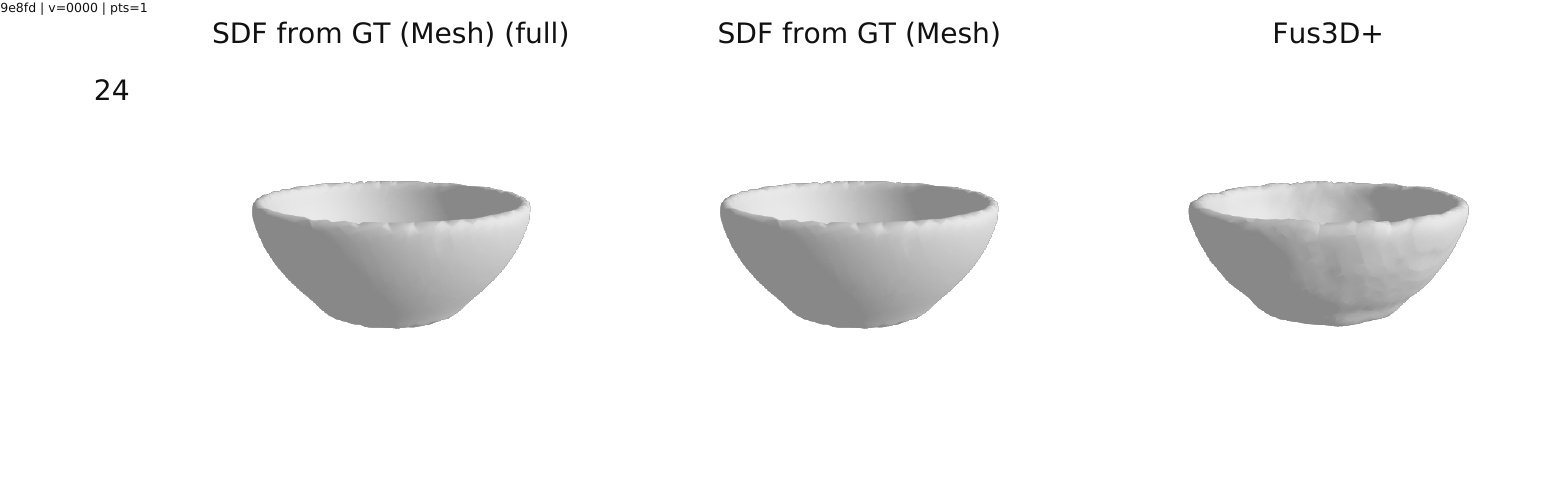}
    \hfill
    \includegraphics[width=0.3\textwidth]{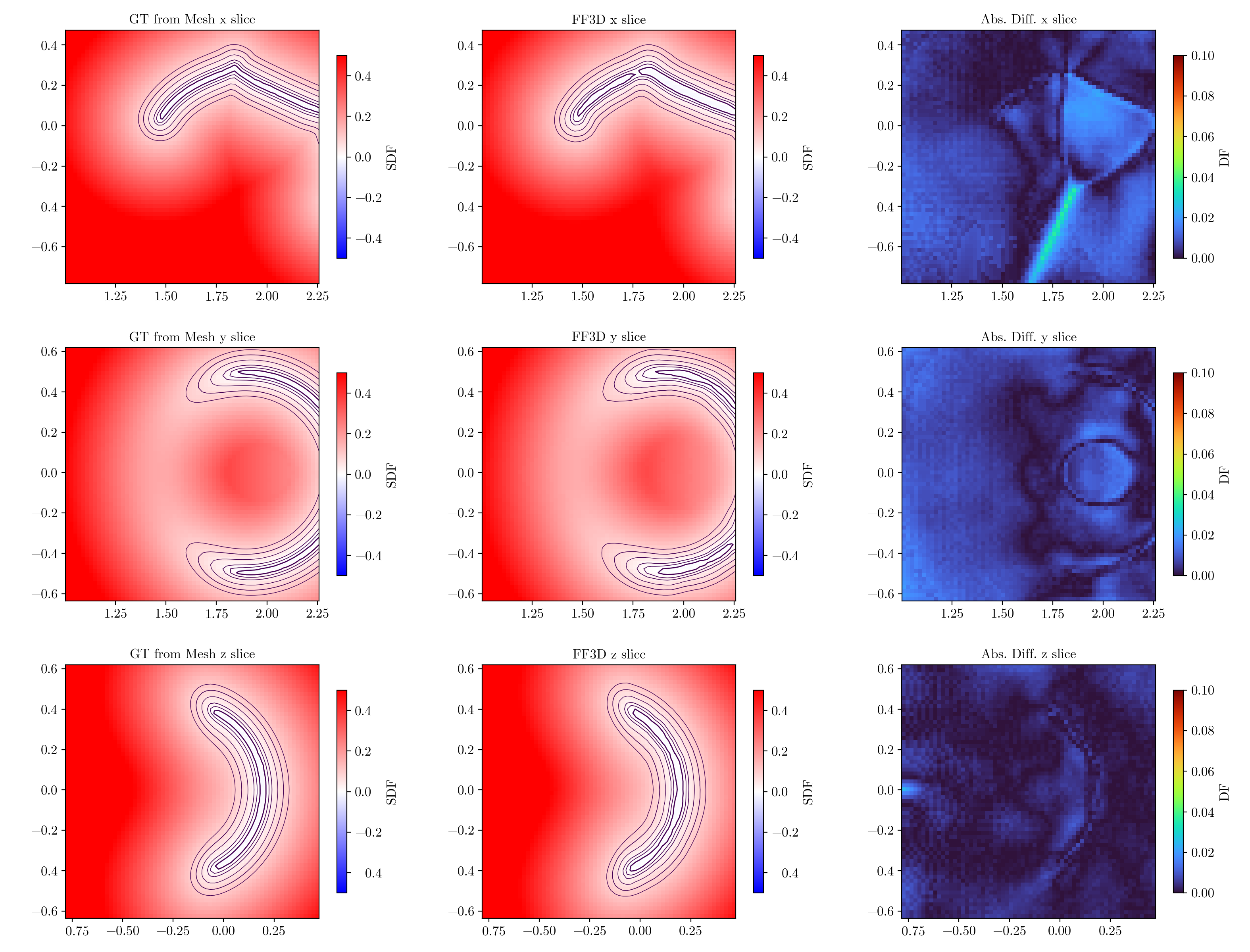}
    \caption{Varying the query domain $\V$, potentially leading to  zoomed or cropped results.  Left: extracted isosurfaces. Right: slices of the SDF volumes.
    \label{fig:variable_query}}
\end{figure}

Our model predicts SDF values for the query volume~$\V$ by conditioning the initial learned embedding~$\latentthreedinit$ with its 3D voxel positions in $\V$.
In the table on the right, we report metrics (bold is better) for a briefly finetuned variant that handles varying query domains (\emph{Fus3D+}), thus allowing for a control mechanism to select the region of interest (see Supp. Fig.~\ref{fig:variable_query} for a visualization).
\emph{Fus3D+ wo. Q} ablates the use of a learned embedding, replacing the initial query entirely by the linearly projected 3D voxel positions.
\emph{Default $\V$} evaluates the default region of interest.
For \emph{Variable $\V$}, a volumetric query with random center and extent was evaluated for every scene.
Notably, the difference between \emph{Fus3D+} and \emph{Fus3D+ wo. Q} is especially pronounced in case of \emph{Variable $\V$}.

\paragraph{Loss Ablation}

\begin{wrapfigure}{r}{0.4\textwidth}
    \vspace{-0.9em}
    \includegraphics[width=\linewidth]{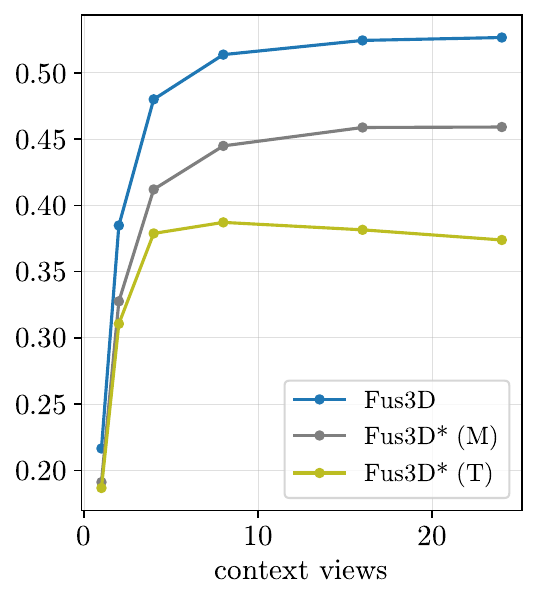}
    \vspace{-2em}
    \caption{Effect of the loss configuration, evaluated with F$_{0.5\eps}$-Score. Fus3D denotes our final model. $*$-variants omit $\gL_\text{SDF}(\nearsurface)$; (M) is trained on mesh-derived SDFs, (T) is trained on TSDFs fused from 24 ground-truth depth maps and camera poses.
    \label{fig:abl_loss}}
    \vspace{-5em}
\end{wrapfigure}

Fig.~\ref{fig:abl_loss} compares different loss configurations, evaluated on our \textsc{Objaverse} test set.
Our final configuration (blue, full loss) consistently outperforms both Fus3D$^*$ variants, showing that the near-surface supervision term $\gL_\text{SDF}(\nearsurface)$ is important for accurate isosurface extraction.
While TSDF-based supervision (T) is potentially more broadly applicable, especially to real-world data, it yields lower surface accuracy than mesh-based supervision (M).
This suggests that the bias inherent to TSDF supervision is transferred to the prediction and may further conflict with Eikonal regularization.

\section{Discussion \& Future Work}
\label{supp:discussion}

While our reconstructions are dense, complete, and plausible even in unobserved regions, fine-scale details can be missing and surfaces may appear over-smoothed.
One contributing factor is the relatively low resolution of our SDF grid.
We deliberately use a straightforward dense decoder to focus model capacity on 3D feature extraction.
Beyond surface reconstruction, the resulting dense 3D features which explicitly represent free space may also be valuable for tasks such as trajectory planning or robotic manipulation.
For high-fidelity surface extraction, however, sparse representations and multi-scale upsampling are a promising direction to improve spatial resolution without prohibitive memory costs.

Another promising direction is to more tightly integrate our extraction module with the backbone, instead of treating the backbone as a ``read-only'' feature provider.
This is conceptually related to spatial-memory approaches, where the model maintains and updates a consistent state as additional views are processed.
Overall, we see higher-resolution (potentially sparse) decoding and a more intertwined backbone--extractor design as complementary avenues to improve detail, robustness, and scalability.

\end{document}